\documentclass{egpubl}

\JournalSubmission    
\copyrightTextTitPag{Preprint.}
\copyrightTextRunPag{Preprint.}

\usepackage[T1]{fontenc}
\usepackage{dfadobe}
\usepackage{graphicx}%
\usepackage{subcaption}
\usepackage{multirow}%
\usepackage{amsmath,amssymb,amsfonts}%
\let\saveproof\proof
\let\saveendproof\endproof
\let\proof\relax
\let\endproof\relax
\usepackage{amsthm}%
\let\proof\saveproof
\let\endproof\saveendproof
\usepackage{mathrsfs}%
\usepackage[title]{appendix}%
\usepackage{textcomp}%
\usepackage{manyfoot}%
\usepackage{booktabs}%
\usepackage{algpseudocode}%
\usepackage{listings}%
\usepackage{array}
\usepackage{tabularx}
\usepackage{stfloats}
\usepackage{url}
\usepackage{verbatim}
\usepackage[table]{xcolor}
\usepackage[ruled,vlined,linesnumbered]{algorithm2e}

\usepackage{balance}
\newcommand{\src}{\mathcal{S}}
\newcommand{\tar}{\mathcal{T}}
\newcommand{\ali}{{\mathbf{A}}}
\newcommand{\fm}{{\mathbf{F}}}
\newcommand{\gm}{{\mathbf{G}}}
\newcommand{\s}{\mathcal{S}}
\newcommand{\st}{\mathcal{T}}
\newcommand{\F}{\mathbf{F}}
\newtheorem{prop}{Proposition}
\usepackage{cite}  
\theoremstyle{thmstyleone}%
\theoremstyle{thmstyletwo}%

\theoremstyle{thmstylethree}%
\BibtexOrBiblatex

\electronicVersion
\PrintedOrElectronic

\usepackage{egweblnk}

\title[EG \LaTeX\ Author Guidelines]%
      {NFR: Neural Feature-Guided Non-Rigid Shape Registration}

\author[Zhangquan Chen \& Puhua Jiang \& Mingze Sun \& Ruqi Huang]
{\parbox{\textwidth}{\centering Zhangquan Chen$^{1}$\footnotemark[1]\orcid{0009-0006-2370-6120}, Puhua Jiang$^{1}$\footnotemark[1]\orcid{0009-0001-1662-0736}, Mingze Sun$^{1}$\thanks{These authors contributed equally to this work.}\orcid{0000-0002-0990-2161} and Ruqi Huang$^{1}$\thanks{Corresponding author: ruqihuang@sz.tsinghua.edu.cn}\orcid{0000-0001-5942-3671}
        }
        \\
{\parbox{\textwidth}{\centering $^1$Tsinghua Shenzhen International Graduate School, Shenzhen, China\\
       }
}
}

\begin{document}

\maketitle
\begin{abstract}
In this paper, we propose a novel learning-based framework for 3D shape registration, which overcomes the challenges of significant non-rigid deformation and partiality undergoing among input shapes, and, remarkably, requires \emph{no} correspondence annotation during training.
Our key insight is to incorporate neural features learned by deep learning-based shape matching networks into an iterative, geometric shape registration pipeline.
The advantage of our approach is two-fold -- On one hand, neural features provide more accurate and semantically meaningful correspondence estimation than spatial features (\emph{e.g., } coordinates), which is critical in the presence of large non-rigid deformations; On the other hand, the correspondences are dynamically updated according to the intermediate registrations and filtered by consistency prior, which prominently robustify the overall pipeline.
Empirical results show that, with as few as dozens of training shapes of limited variability, our pipeline achieves state-of-the-art results on several benchmarks of non-rigid point cloud matching and partial shape matching across varying settings, but also delivers high-quality correspondences between unseen challenging shape pairs that undergo both significant extrinsic and intrinsic deformations, in which case neither traditional registration methods nor intrinsic methods work.
Our code is available at \url{https://github.com/rqhuang88/NFR}.


\end{abstract}
\section{Introduction}\label{sec:intro}
Estimating dense correspondences between 3D human shapes is pivotal for a multitude of human-centric computer vision applications, including 3D reconstruction~\cite{yu2018doublefusion}, 3D pose estimation~\cite{pe}, and animation~\cite{paravati2016point} among others. In this paper, we tackle the challenging task of estimating correspondences between \emph{unstructured} point clouds sampled from surfaces undergoing significant non-rigid deformations, irrespective of whether the point clouds are \emph{full} or \emph{partial}.

In fact, the shape matching task has garnered increasing attention within the community~\cite{zeng2021corrnet3d, lang2021dpc, lie, nie, cao2023}.
These methods predominantly adopt a data-driven approach to learn embedding schemes that project point clouds into high-dimensional spaces. 
By elevating the point clouds to these high-dimensional spaces, non-rigid deformations are more effectively characterized compared to the ambient space, $\mathbb{R}^3$. The optimal transformations, and consequently the correspondences, are then estimated within these embedded spaces. Despite the great progresses, such approaches are hindered by several limitations: 1) Their performance on previously unseen shapes is largely unpredictable; 2) The learned high-dimensional embeddings lack intuitive geometric interpretation, making the resulting correspondences challenging to evaluate and analyze in the absence of ground-truth labels; 3) There is still a large room for improvement in estimating dense correspondences between human shapes with \emph{partiality}.

\begin{figure}[!t]
    \centering
    \includegraphics[width=0.5\textwidth]{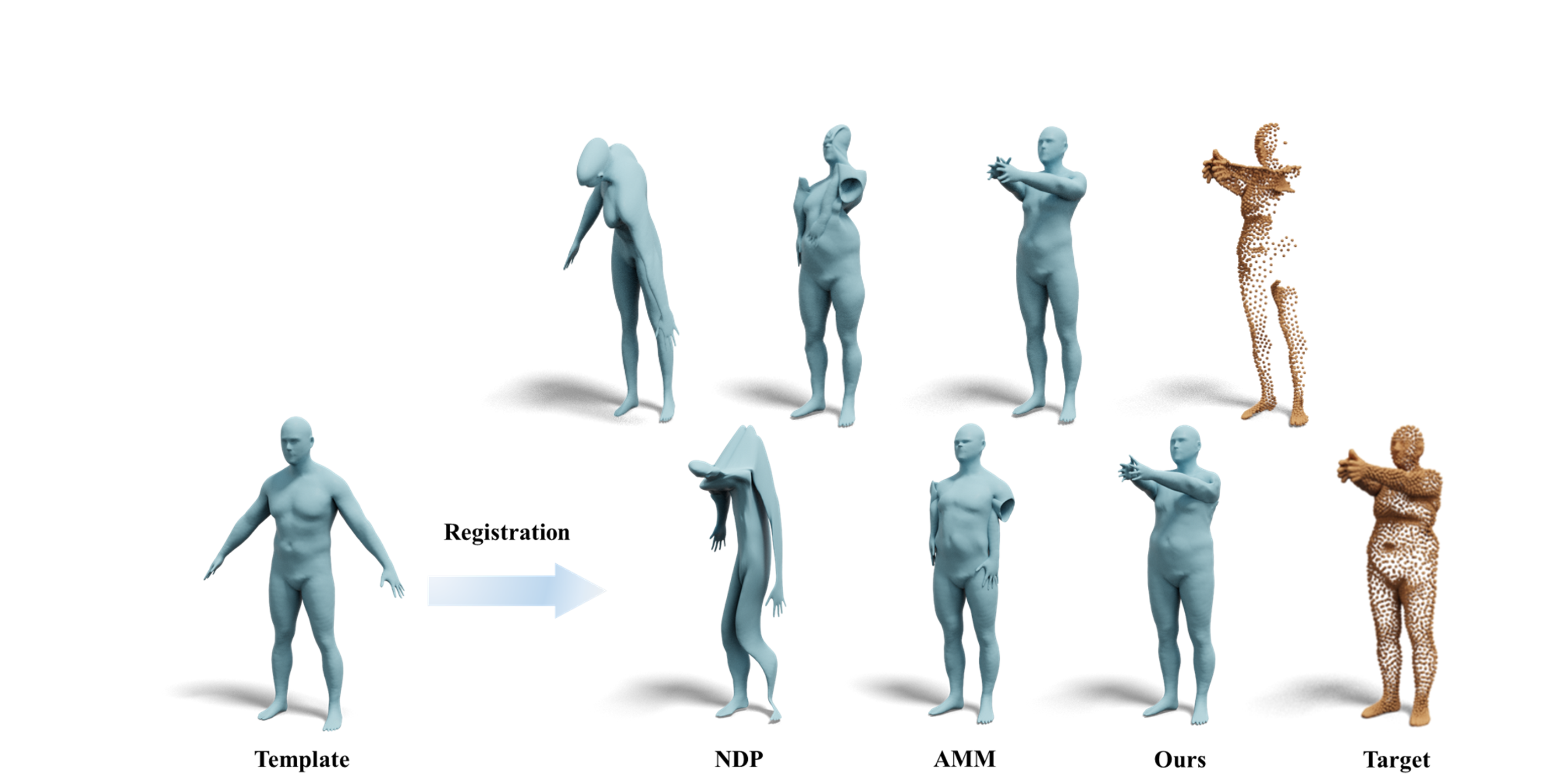}
    \caption{Shape registration methods like NDP~\cite{li2022non} and AMM~\cite{AMM} estimate intermediate correspondences via extrinsic proximity, therefore suffering from large intrinsic deformations. In contrast, our method successfully deforms a FAUST template (the left-most mesh) to another individual of a different pose (the right-most full or partial point cloud). }
    \label{fig:row}
    \vspace{-2em}
\end{figure}

On the other hand, shape registration itself does not serve as a direct rescue to our problem of interest, as the previous approaches typically rely on the premise that the undergoing non-rigid deformation can be approximated by a set of local, small-to-moderate, rigid deformations, which severely hinders their performance in the presence of large deformations (see Fig.~\ref{fig:row}) and heterogeneity.

Motivated by the above observations, we propose a \emph{neural feature-guided framework} (NFR) that synergistically combines the strengths of classic shape registration and learning-based embedding techniques. In a nutshell, we leverage the estimated correspondence from the latter to guide shape deformation via the former iteratively. 

Our key insight is to enforce similarity between the deformed source and target in \emph{both the ambient space and the learned high-dimensional space}. Intuitively, in the presence of large deformations, the correspondences learned in the high-dimensional space are more reliable than those based solely on proximity in the ambient space. Conversely, accurately deforming the source mesh to align with the target point cloud enhances spatial similarity in the ambient space, which in turn increases similarity in the embedded space, leading to more precise correspondences. Ultimately, this approach enables us to compute correspondences between raw point clouds using a shared source mesh as a central hub. As demonstrated in Sec.~\ref{sec:exp}, our method allows for the selection of the source shape during inference, which can be independent of the training data. While conceptually straightforward, we aim to enhance the performance, robustness, and practical utility of our pipeline through several tailored designs. First and foremost, the key component of our pipeline is an embedding scheme for accurately and efficiently estimating correspondences between the deformed source shape and the target \emph{point cloud}. Specifically, to leverage the advantages of Deep Functional Maps (DFM)~\cite{li2022attentivefmaps, cao2022, dual}, the current state-of-the-art approach for matching \emph{triangular meshes}, we pre-train an unsupervised DFM~\cite{dual} on meshes as a teacher network. We then train a point-based feature extractor (i.e., an embedding scheme for points) as a student network on the corresponding vertex sets using natural self-supervision. Unlike the approach in~\cite{cao2023}, which heavily relies on DiffusionNet~\cite{diffusionNet} to extract intricate structural details (e.g., Laplace-Beltrami operators) from both mesh and point cloud inputs, the teacher-student paradigm allows us to utilize a more streamlined backbone—DGCNN~\cite{dgcnn}. Secondly, in contrast to prior works that either implicitly~\cite{li2022non, AMM} or explicitly~\cite{sharma20, cao2023, nie} require rigidly aligned shapes for input/initialization, or demand dense correspondence labels~\cite{lie}, we train an orientation regressor on a large set of synthetic, rigidly aligned shapes to \emph{automatically} pre-process input point clouds in arbitrary orientations. Lastly, we introduce several innovative components in our pipeline, including a dynamic correspondence updating scheme, bijectivity-based correspondence filtering, and a two-stage registration process.

Moreover, we supplement the above framework for estimating correspondences between \emph{partial point clouds and complete source shapes}. In particular, we assume to be given a set of \emph{complete meshes} as training data. During training, we randomly pick a pair of meshes $\s, \st$, and sample on $\st$ for a partial point cloud $\st_p$, without preserving any mesh structure. Recall that, in DPFM~\cite{attaiki2021dpfm}, one first constructs a graph-like structure on $\st_p$ and then computes spectral embedding on it, which is both slow and unstable. In contrast to that, we consider to assign the spectral embedding of $\st$ to its subset $\st_p$ via the natural inclusion map, forming a novel \emph{spatially truncated spectral embedding} with respect to $\st_p$ and $\st$. As shown in Prop.~\ref{prop} in Sec.~\ref{sec:method}, under certain conditions, we prove that the transformation between the spectral embedding of $\s$ and that of $\st_p$ can always be set as the functional map from $\s$ to $\st$, \emph{regardless} of how $\st_p$ is generated. The above procedure leads to a simple yet effective training scheme. Namely, we can use arbitary point-based feature extractor to obtain per-point feature on both $\st_p$ and $\s$, compute point-wise map $T:\st_p \rightarrow \s$ via proximities in the feature space, and encode $T$ into transformation between the spectral embedding of $\st_p$ and $\s$. Finally, the functional map from $\s$ to $\st$ can be used as the supervision signal on the obtained transformation. One last missing puzzle from the above pipeline is the source of functional maps among the full shapes. In fact, such can be either given a prior (\emph{e.g.}, when we train on a set of synthetic shapes~\cite{SMPL_2015}) or learned (\emph{e.g.}, with the SOTA DFM~\cite{dual}). Last but not least, to further improve performance, we incorporate our learned point feature extractor with a neural enhanced shape registration framework, as described above, which applies a geometric optimization during inference. 

Overall, our framework enjoys the following advantages: 1) Due to the hybrid nature, our framework can effectively manage point clouds that undergo significant deformation and/or exhibit heterogeneity; 2) Our feature extractor, being self-supervised by a deep functional maps network, eliminates the need for correspondence annotation throughout the process; 3) The core operation is conducted in the ambient space, allowing for more efficient, accurate, and straightforward analysis of registration/mapping results compared to purely learning-based approaches; 4) Utilizing a data-driven orientation regressor, we have developed an automatic pipeline for estimating correspondences between deformable point clouds; 5) We introduce a novel self-supervised training scheme for estimating correspondences between partial point clouds and complete source shapes, along with a robust partial-to-full shape registration pipeline.

\begin{figure*}[!t]
    \centering
    \includegraphics[width=\textwidth]{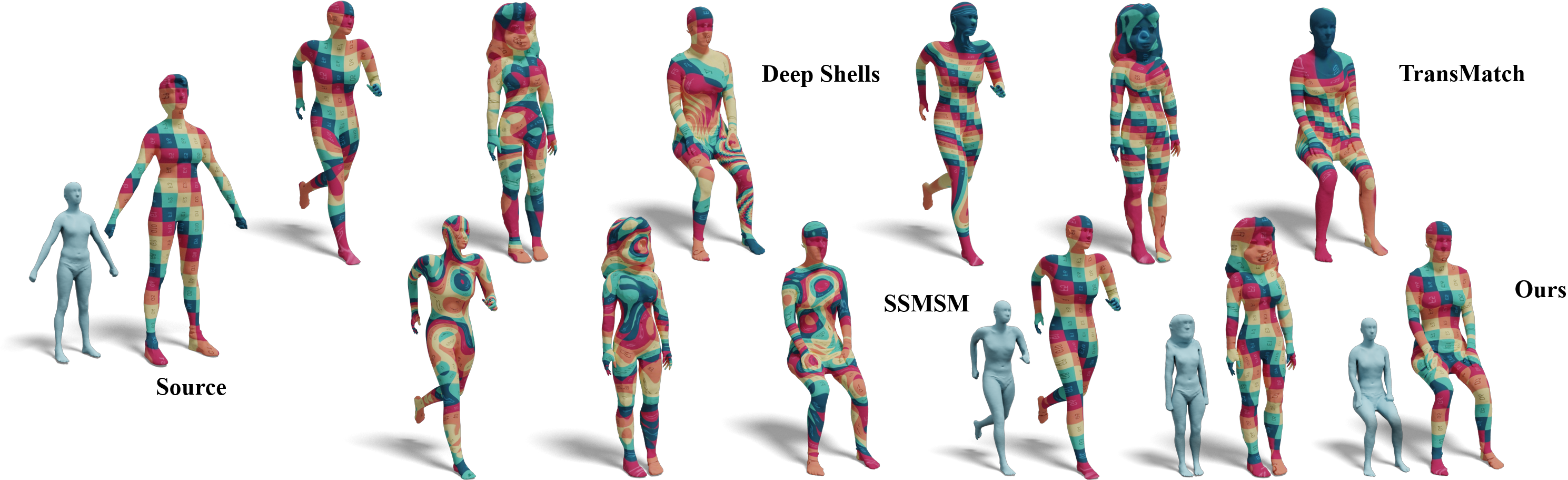}
    \caption{We estimate correspondences between heterogeneous shapes from SHREC'07 with four learning-based methods, all trained on the SCAPE\_r dataset. Our method outperforms the competing methods by a large margin. Remarkably, our method manages to deform a SCAPE template shape to heterogeneous shapes, as indicated by the blue shapes.}
    \label{fig:teaser}
    \vspace{-0.8em}
\end{figure*}

\begin{figure*}[!t]
    \centering
    \includegraphics[width=\textwidth]{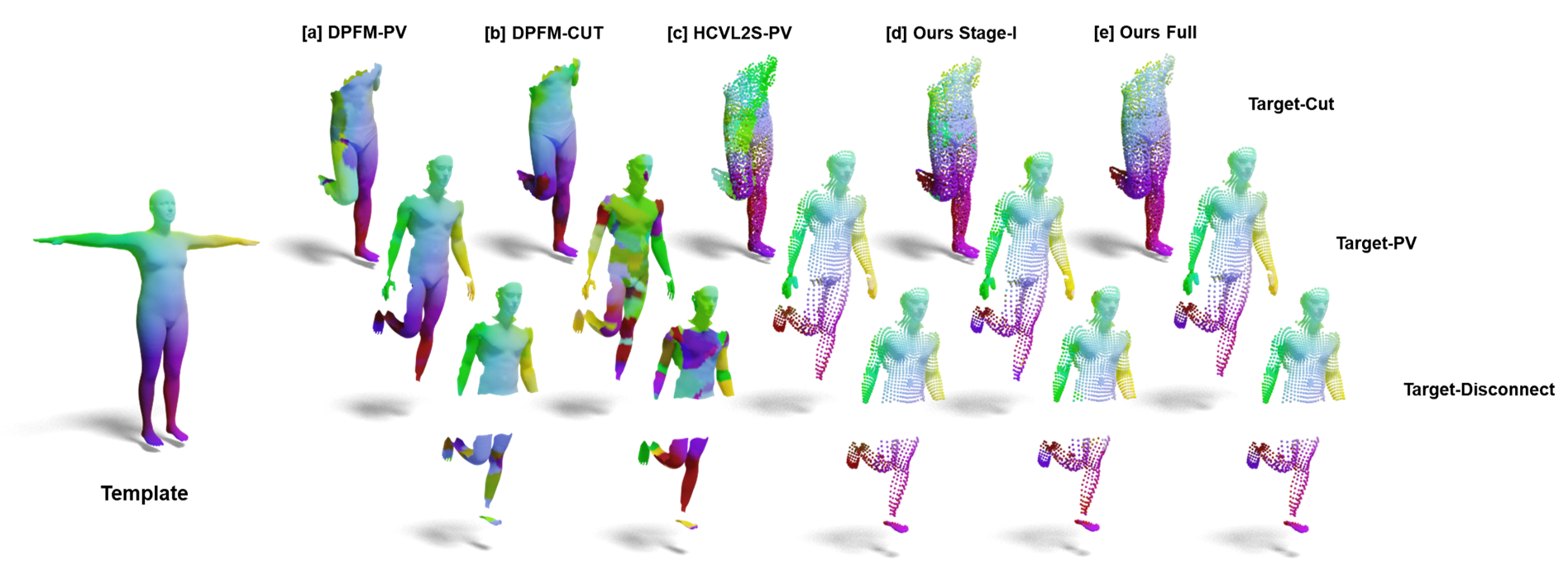}
    \caption{Matching different partial shapes to a common template shape (left-most). (a) DPFM~\cite{attaiki2021dpfm} trained on \textbf{Ours-S\&F} dataset; (b) DPFM~\cite{attaiki2021dpfm} trained on \textbf{CUT}~\cite{attaiki2021dpfm}; (c) HCLV2S~\cite{huang20} trained on large scale \textbf{SURREAL}~\cite{huang20} dataset ; (d) Ours trained on \textbf{Ours-S\&F} dataset with only stage-I; (e) Ours trained on \textbf{Ours-S\&F} dataset with stage-I and stage-II. DPFM~\cite{attaiki2021dpfm}   and HCVL2S~\cite{huang20} fails to handle unseen partial shapes effectively. Remarkablely, DPFM significantly struggles with disconnected shapes. Our method outperforms the competing methods by a large margin and achieve the consist results across different partiality.  }
    \label{fig:teaser_partial}
    \vspace{-2em}
\end{figure*}

As illustrated in Fig.~\ref{fig:teaser}, our framework, trained on the SCAPE~\cite{scape} dataset, demonstrates excellent generalization capabilities and significantly outperforms competing methods on the challenging SHREC07-H dataset, which exhibits substantial variability in both extrinsic orientation and intrinsic geometry. Additionally, Fig.~\ref{fig:teaser_partial}(d) shows that our feature extractor already provides robust and reasonable correspondences across different partiality patterns. The extra refinement step further addresses local mismatches, resulting in satisfying outcomes (Fig.~\ref{fig:teaser_partial}(e)). We further emphasize the importance of generalizability in partial shape matching. Unlike full shapes, one can theoretically generate an infinite number of partial subsets from a given surface, with varying degrees of connectedness, partiality patterns, and sizes. As shown in columns (a-c) of Fig.~\ref{fig:teaser_partial}, recent advances such as DPFM~\cite{attaiki2021dpfm} and HCLV2S~\cite{huang20} fail to generalize to unseen partiality patterns. In contrast, our pipeline exhibits strong generalizability, even in practical partial human shape-matching tasks.
\section{Related Works}
\label{sec:related}


\subsection{Non-rigid Shape Registration} 
Non-rigidly aligning shapes presents greater complexity compared to the rigid counterpart due to the intricate nature of deformation models. In general, axiomatic approaches~\cite{nicp} assume the deformation of interest can be approximated by local, small-to-moderate, rigid deformations, therefore suffer from large intrinsic deformations. Recent advances, including~\cite{fgr, li2018robust, AMM}, have primarily aimed at enhancing efficiency and robustness against noise and outliers. Concurrently, there is a growing interest in integrating deep learning techniques into non-rigid shape registration, as seen in works like~\cite{bozic2020deepdeform, bozic2020neural, huang_multiway_2022}. Among them, NDP~\cite{li2022non} is similar to our method in the sense that the non-learned version follows an optimization-based approach based on some neural encoding. However, its registration process is solely driven by proximity in the ambient space, which compromises the effectiveness for our specific problem. Perhaps the most relevant approach with our method along this line is TransMatch~\cite{trappolini2021shape}, which utilizes a supervised learning framework to train a transformer for predicting 3D flows directly from input point clouds. As demonstrated in Fig.\ref{fig:teaser} and discussed further in Sec.\ref{sec:exp}, TransMatch generalizes poorly to unseen shapes. 

\subsection{Non-rigid Shape Matching} is a long-standing problem in computer vision and graphics.  Unlike the rigid counterpart, non-rigidly aligning shapes is more
challenging owing to the complexity inherent in deformation models. Generally, axiomatic approaches ~\cite{nicp} assume the deformation of interest can be approximated by local, small-to-moderate, rigid deformations, therefore suffer from large intrinsic deformations. Recent advancements~\cite{fgr,li2018robust,} have primarily focused on enhancing the efficiency and robustness of methods in the face of noise and outliers. Simultaneously, there's a growing trend towards integrating deep learning techniques~\cite{bozic2020deepdeform, bozic2020neural, huang_multiway_2022}. Contrary to the methods mentioned earlier, several recent approaches directly establish correspondences between pairs of point clouds~\cite{lang2021dpc,li2022lepard,zeng2021corrnet3d}. Generally, these methods involve embedding point clouds into a canonical feature space and then estimating correspondences based on proximity in this space. Since intrinsic information is not explicitly formulated in these methods, they can suffer from significant intrinsic deformations and often generalize poorly to unseen shapes.




\subsection{Functional Maps}
Based on the spectral embedding, \emph{i.e., }the truncated eigenbasis of the Laplace-Beltrami operator~\cite{pinkall1993computing} on a shape, functional maps~\cite{ovsjanikov2012functional} encode point-wise maps into linear transformation between the respective spectral spaces. This approach effectively transforms the problem of finding point-wise correspondences into a simpler task of aligning spectral coefficients. Originating from the foundational work on functional maps, along with a series of follow-ups~\cite{nogneng2017informative, huang2017adjoint, BCICP, zoomout, huang2020consistent}, spectral methods have made significant  progress in addressing the axiomatic non-rigid shape matching problem. 

Unlike axiomatic functional maps frameworks, which focus on optimizing correspondences represented in a spectral basis (referred to as functional maps) \emph{with hand-crafted features as a prior}, DFM~\cite{litany2017deep} adopts an inverse viewpoint. It aims to search for the optimal features such that the induced functional maps satisfy certain structural priors as well as possible.

However, because of the heavy dependence of eigenbasis of Laplace-Beltrami operators, DFM is primarily designed for shapes represented by triangle meshes and can suffer notable performance drop when applied to point clouds without adaptation~\cite{cao2023}. In fact, inspired by the success of DFM, several approaches~\cite{nie, cao2023,dfr} have been proposed to leverage intrinsic geometry information carried by meshes in the training of feature extractors tailored for non-structural point clouds.

While significant advancements have been made in full shape matching, there remains considerable room for improvement in estimating dense correspondences between shapes with partiality. Functional maps representation~\cite{rodola2016partial, rodola2017partial} has already been applied to partial shapes. However, the partial functional maps (PFM) formulation, as proposed in~\cite{rodola2016partial} has its limitations. Building on this foundation, recent learning-based frameworks such as DPFM~\cite{attaiki2021dpfm} and SSMSM~\cite{cao2023} have shown notable improvements over previous axiomatic approaches. However, both axiomatic and learning-based lines of work typically assume the input to be a \emph{connected mesh}, with the exception of~\cite{cao2023}, which relies on graph Laplacian construction~\cite{sharp2020laplacian} in its preprocessing.

More recently, several works have advanced partial shape matching on meshes.
Ehm~\emph{et al.}~\cite{ehm2024partial} propose a geometric consistency-based approach for partial-to-partial mesh matching, while EchoMatch~\cite{xie2025echomatch} introduces correspondence reflection for partial-to-partial matching.
NAM~\cite{vigano2025nam} refines mesh correspondences via neural adjoint maps.
However, these methods operate under the assumption of mesh input with known connectivity, which enables construction of Laplace-Beltrami operators.
In contrast, our method is specifically designed for unstructured point clouds and combines learned features with geometric registration, addressing a fundamentally different and arguably more challenging setting.


\subsection{Learning-based Human Partial Shape Matching}
When matching partial human scan, a popular approach is to perform template matching. Early works, such as ~\cite{yu2018doublefusion,yu2017bodyfusion}, explicitly modeled shape deformations and employed extrinsic distance metrics for matching purposes. 
While some methods~\cite{lin2022occlusionfusion} utilize deep learning to predict optical flow for guiding registration, these methods are still only effective in cases of minor deformations.
More recent methods~\cite{groueix20183d,hanocka2018alignet} have shifted towards learning models of shapes. They formulate  template matching as an optimization of parameters within a learned latent space. The authors of ~\cite{huang20} suggest a different approach: formulating the computation of dense correspondences as the initialization and synchronization of local transformations between the scan and the template model. However, these methods often suffer from poor generalizability to unseen shapes. 
\section{Methodology}
\label{sec:method}

Given a pair of shapes $\src, \tar$, our target is to deform $\src$ to non-rigidly align with $\tar$. We assume that $\src$ is represented as a triangle mesh so that we can effectively regularize the deformed shape by preserving local intrinsic geometry. On the other hand, we require \emph{no} structural information on $\tar$ and generally assume it to be a \emph{point cloud}. Our pipeline can also be extended to compute correspondences between two raw point clouds $\tar_1, \tar_2$. To this end, we fix a template mesh $\src$, perform respective shape registration between $\src$ and the target point clouds, and finally compute the map by composition $T_{12} = T_{s2}\circ T_{1s}$. Specifically, the non-rigid registration approach developed to tackle the problem of \emph{full-2-full} is named \textbf{DFR}, and the methodology designed for \emph{partial-2-full} problem is named \textbf{Partial-DFR}. 

\subsection{DFR}
\begin{figure}[!t]
    \centering
    \includegraphics[width=0.5\textwidth]{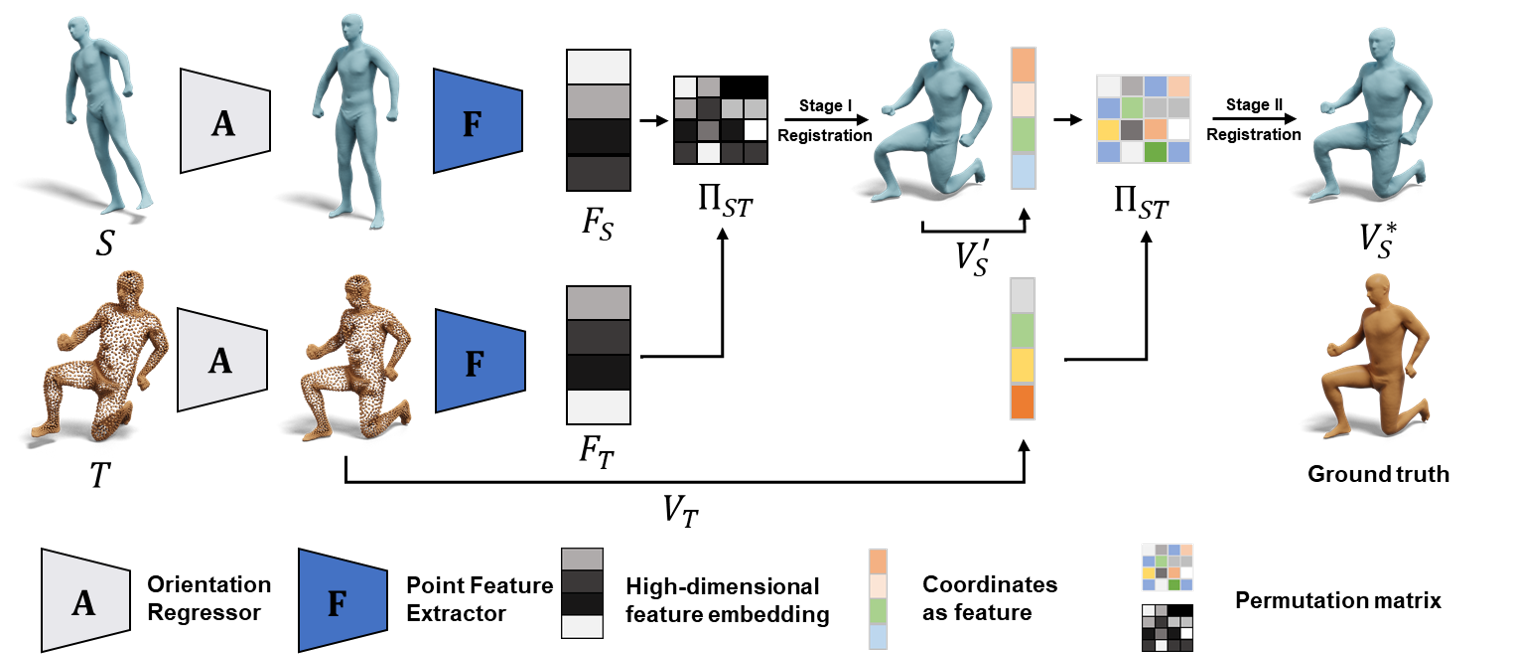}
    \caption{The schematic illustration of our pipeline. $\ali$ is a pre-trained orientation regressor for aligning input shapes. Then a pre-trained feature extractor $\fm$ embeds them into a high-dimensional canonical space. During the iterative optimization procedure of registration, correspondences are dynamically updated according to learned features (Stage-I) and coordinates (Stage-II) of the intermediate shapes. See more details in the text. 
    }\label{fig:pipeline}
    \vspace{-0.8em}
\end{figure}
In this subsection, we give a brief recap of the DFR introduced in our previous work~\cite{dfr}.

Our DFR registration pipeline is shown in Fig.~\ref{fig:pipeline}, which consists of three main components: 1) An orientation regressor, $\ali$, for \emph{extrinsically} aligning input shapes, either mesh or point cloud; 2) A point feature extractor, $\fm$, trained under deep functional maps scheme; 3) A registration module that iteratively optimizes for deformations non-rigidly aligning $\src$ with $\tar$. In particular, it takes the rigidly aligned shapes from 1) as input, and leverages 2) to update correspondences during optimization.  

Though our pipeline leverages a pre-trained feature extractor as registration prior, we highlight that neither the source nor the target is necessarily within or close to the respective training set. We provide throughout experimental evidence in Sec.~\ref{sec:exp} showing the generalizability of our pipeline. 

\subsubsection{Orientation Regressor}\label{sec:ori}
Our first objective is to align input point clouds with arbitrary orientations into a canonical frame. We take a data-driven approach by training an orientation regressor~\cite{chen2022projective} on $5000$ synthetic SURREAL shapes from~\cite{varol17_surreal}, which are implicitly aligned to a canonical frame by the corresponding generative codes. 

\subsubsection{Point Feature Extractor}\label{sec:dfm}
In order to efficiently and accurately estimate correspondences between the deformed source shape and target point cloud during registration, our next goal is to train a feature extractor for point clouds, which is intrinsic-geometry aware. The authors of~\cite{cao2023} propose a multi-modal feature extractor based on DiffusionNet~\cite{diffusionNet}, which can process point clouds with an extra step of \emph{graph} Laplacian construction~\cite{sharp2020laplacian}.
Though it has demonstrated satisfying performance accommodating point cloud representation in~\cite{cao2023}, the explicit graph Laplacian construction is computationally heavy. Hence, we adopt the modified DGCNN proposed in~\cite{nie} as our backbone, which is lightweight and robust regarding the sampling density and sizes of point clouds.

In particular, our training follows a teacher-student paradigm. Namely, we first train a deep functional maps (DFM) network, $\gm$, on a collection of meshes. Then we train a DFM network $\fm$ on the corresponding vertex sets, with an extra self-supervision according to the inclusion map between meshes and their vertex sets. In other words, the feature produced by $\fm$ is \emph{point-wisely} aligned with that produced by $\gm$. 

\subsubsection{Total cost function}
The total cost function $E_{\text {total }}$ combines the terms with the weighting factors $\lambda_{c d}, \lambda_{c o rr}$, and $\lambda_{arap}$ to balance them:
\begin{equation}\label{eqn:total}
    E_{\mbox{total}}=\lambda_{\mbox{cd}} E_{\mbox{cd}}+\lambda_{\mbox{corr}} E_{{\mbox{corr}}}+\lambda_{\mbox{arap}} E_{\mbox{arap}}
\end{equation}



We describe our algorithm for minimizing $E_{\mbox{total}}$, which is shown in Alg.~\ref{alg:algorithm1}

\subsubsection{Two-stage Registration}
Finally, we observe that solely depending on learned features to infer correspondence is sub-optimal. At the converging point, the deformed source shape is often at the right pose but has deficiencies in shape style. To this end, we perform a second-stage registration, based on the coordinates of deforming source and target. As shown in Sec.~\ref{sec:abl}, such a design is clearly beneficial. 

\begin{algorithm}[!t]
        \relsize{-1}
	\caption{Shape registration pipeline.}
	\label{alg:algorithm1}
	\KwIn{Source mesh $\src = \{ \mathcal{V} , \mathcal{E}  \}$ and target point cloud $\tar$;Trained point feature extractor \textbf{F}}
	\KwOut{ $\textbf{X} ^*$ converging to a local minimum of $E_{\mbox{total}}$; Deformed source model $ \{ \mathcal{V}^* , \mathcal{E}  \}$; Correspondence $\Pi_{\mathcal{S}\mathcal{T}}^*,\Pi_{\mathcal{T}\mathcal{S}}^*$ between $\src$ and $\tar$.}  
	\BlankLine
	Initialize deformation graph $\mathcal{DG}$  and  $\textbf{X} ^{(0)}$ with identity transformations;
     $F_{\mathcal{T}} = \textbf{F}(\tar)$;k = 0;

	\While{\textnormal{True}}{
        Update source vertex $ \mathcal{V}^{(k)} $ by Eqn.(\ref{eqn:dg});
         
        \If{$k \% 100 == 0$ and Flag == Stage-I }{
        $F^{(k)}_{\mathcal{S}} = \textbf{F}(\mathcal{V}^{(k)})$;
        $\Pi_{\src\tar}^{(k)}= \textbf{NN}(F^{(k)}_{\src},F_{\tar} ) $ ; $\Pi_{\tar\src}^{(k)}= \textbf{NN}(F_{\tar},F^{(k)}_{\src} ) $;
        }

        \If{$k \% 100 == 0$ and Flag == Stage-II }{
        $\Pi_{\src\tar}^{(k)}= \textbf{NN}(\mathcal{V}^{(k)},\tar ) $ ; $\Pi_{\tar\src}^{(k)}= \textbf{NN}(\tar ,\mathcal{V}^{(k)} ) $;
        }

        Compute the set of filtered correspondences $\mathcal{C}^k$ ;

        $\textbf{X}^{(k+1)} =  \arg\min{E_{\mbox{total}}}$\text{ by Eqn.(\ref{eqn:total})};
        
         \textbf{if} converged and Flag == Stage-I \textbf{then} Flag = Stage-II;
         
        \textbf{if} converged and  Flag == Stage-II \textbf{then} 
            \textbf{return }$\Pi_{\mathcal{S}\mathcal{T}}^{*}$;$\Pi_{\mathcal{T}\mathcal{S}}^{*}$; $ \mathcal{V}^{*}$; 

         k = k + 1;
    }
\end{algorithm}

\subsection{Partial-DFR}
In the forthcoming section, we delve into the registration of partial cases via deep functional maps prior, namely Partial-DFR. We start by giving a theoretical analysis on transformations between spectral embeddings of full shapes and that of partial shapes sampled from the former (Sec.~\ref{sec:theory}).
Then based on our theoretical result, we propose a self-supervised scheme for learning an intrinsic-geometry-aware point-based feature extractor on a set of meshes (Sec.~\ref{sec:learning}).
Finally, we show how our point-based feature extractor can be integrated to the recent neural enhanced shape registration framework~\cite{dfr} (Sec.~\ref{sec:reg}).

\subsubsection{Partial Theoretical Analysis}\label{sec:theory}
\begin{figure}[]
    \centering
    \includegraphics[width=0.4\textwidth]{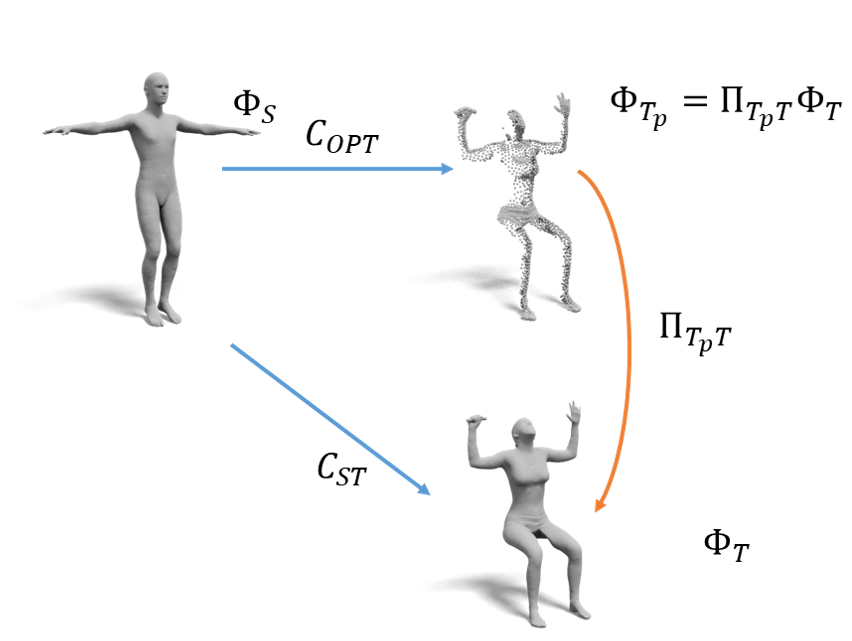}
    \caption{Illustration of the partial spectral embedding construction and partial to full functional map processing during training.}
    \label{fig:partial_opt}
    \vspace{-2em}
\end{figure}

In this section, we further analyze the adaptability of DFR under partial cases. The difference between this and DFR is shown in Fig.~\ref{fig:partial_opt}. We let $\Phi_\s, \Phi_\st\in \mathcal{R}^{n\times k}$ be the first $k$ eigenvectors of the respective cotangent Laplacian~\cite{pinkall1993computing}, and $\Delta_\s, \Delta_\st \in \mathbb{R}^{k\times k}$ be the diagonal matrices storing the corresponding eigenvalues in the diagonals. 
The functional map, $C_{\s\st}$, from $\s$ to $\st$ is then defined as:
\begin{equation}\label{eqn:fmap}
   C_{\s\st} = \Phi_\st^{\dagger}  \Pi_{\st \s} \Phi_\s, 
\end{equation}
where $^{\dagger}$ denotes pseudo inverse of a matrix. 
We follow the perspective in~\cite{zoomout, limitshape} and recognize $C_{\s\st}$ as a transformation \emph{aligning} $\Phi_\s$ and $\Phi_\st$. 
It is pointed out in~\cite{zoomout, limitshape} that
\begin{equation}\label{eqn:fmap2}
    C_{\s\st} = \arg\min_{C} \Vert \Phi_\st C - \Pi_{\st \s} \Phi_\s \Vert_{\mbox{Fro}}^2. 
\end{equation}
In other words, $C_{\s\st}$ defined in Eqn.(\ref{eqn:fmap}) best aligns the spectral embeddings with respect to a given point-wise map $\Pi_{\st \s}$.
In particular, if $\Pi_{\s\st}$ is a isometry between $\s$ and $\st$, then the induced functional map $C_{\s\st}$ enjoys nice structural properties~\cite{ovsjanikov2012functional} including 1) Orthogonality: $C_{\s\st}^TC_{\s\st} = I_k$, where $I_k$ is the $k\times k$ identity matrix; 2) Commutativity with Laplacian operators: $C_{\s\st} \Delta_\st = \Delta_\s C_{\s\st}$. 

Now, we consider $\st_p$, a set of $n_p$ vertices sampled from $\st$. 
We denote the natural inclusion map as $\Pi_{\st_p\st}\in \mathbb{R}^{n_p\times n}: \st_p \rightarrow \st$, which leads to a composed map $\Pi_{\st_p\s} = \Pi_{\st_p\st}\Pi_{\st \s}: \st_p \rightarrow \s$. 
In the following, we focus on the functional map $C_{\s \st_p}$, which is associated with $\Pi_{\st_p\s}$, namely, the partial-to-full map of interest. 

If we follow the framework of partial functional maps~\cite{rodola2016partial}, then we need to first \emph{compute} the eigenbasis of $\st_p$, which further requires meshing, or at least graph construction of top of the discrete sampling points. 
However, in practice, it is generally difficult to guarantee the connectedness of $\st_p$. 
When $\st_p$ contains multiple distant disconnected components, we have to deal with them one by one, resulting in multiple functional maps, which are heavy but also hard to ensure global consistency. 

In contrast, we propose to consider the functional maps encoded in the \emph{spatially truncated spectral embedding} for $\st_p$. 
Specifically, we set 
\begin{equation}\label{eqn:steigen}
    \Phi_{\st_p} = \Pi_{\st_p\st} \Phi_\st    
\end{equation}

\begin{prop}\label{prop}
    Let $\s, \st$ be a pair of shapes each having non-repeating Laplacian eigenvalues, which are the same (\emph{i.e., } $\Delta_\s = \Delta_\st$), and $\Pi_{\st\s}$ be an isometry between $\st$ and $\s$. $\Phi_{\s}, \Phi_{\st}$ are the corresponding eigenvectors. 
    On the other hand, we let $\st_p$ be a sub-sampled set of vertices of $\st$, and $\Phi_{\st_p}$ be the spatially truncated embedding defined in Eqn.~(\ref{eqn:steigen}), then $C_{\s\st}$ defined in Eqn.(~\ref{eqn:fmap}) satisfies: 
    \begin{equation}\label{eqn:prop}
        C_{\s\st} =  \arg\min_{C} \Vert \Phi_{\st_p} C - \Pi_{\st_p \s} \Phi_\s \Vert_{\mbox{Fro}}^2.
    \end{equation}
\end{prop}
Proposition~\ref{prop} essentially argues that, under certain conditions, the full-to-full functional map $C_{\s\st}$ exactly aligns $\Phi_{\st_p}$ to $\Phi_{\s}$.

\noindent\textbf{Why not apply DFR directly to partial inputs?}
A natural question is whether the standard DFR pipeline can be directly applied to partial-to-full matching without the proposed Partial-DFR modifications.
We argue that this is inadequate for two key reasons.
First, DFR employs \emph{bidirectional} Chamfer distance as a surface fitting term (Sec.~\ref{sec:reg}), which enforces proximity in both directions between the template and target. When the target is a partial point cloud, the template-to-target direction forces the \emph{entire} template to collapse toward the visible region, leading to severe geometric distortion.
Second, the feature extractor trained under DFR assumes full-to-full correspondence, producing unreliable features when applied to partial inputs.
As we show in Tab.~\ref{table:dfr} (Sec.~\ref{sec:abl}), the DFR feature extractor yields a geodesic error of $47.29$ on the PFARM benchmark (\emph{vs.}~$18.41$ for our Partial-DFR feature extractor), confirming the necessity of our dedicated partial training scheme.
In contrast, our Partial-DFR pipeline replaces the bidirectional CD with a \emph{unidirectional} variant and leverages the spatially truncated spectral embedding established above, which avoids both the graph Laplacian construction on disconnected partial point clouds and the template collapse issue.

\subsubsection{Self-Supervised Partial Training Scheme}\label{sec:learning}
\begin{figure}[!hb]
    \centering
    \includegraphics[width=0.5\textwidth]{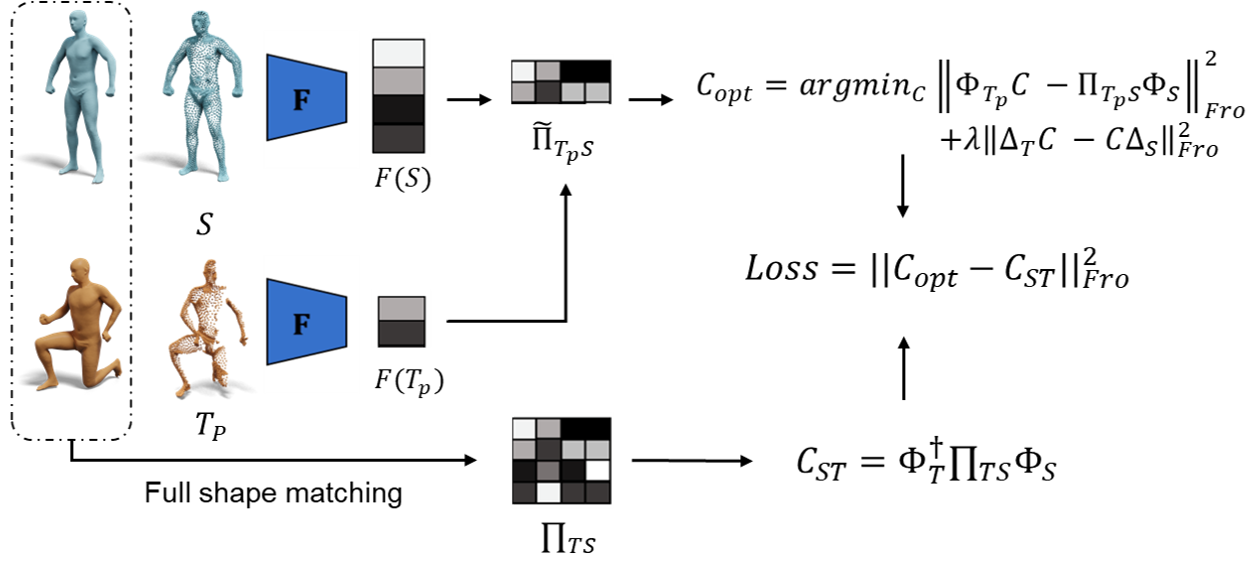}
    \caption{Overview of our self-supervised partial shape scheme for learning an intrinsic-geometry-aware point feature extractor.    }\label{fig:trainpipeline}
\end{figure}

In order to efficiently and accurately estimate correspondences between a source shape
and a target partial point cloud, we propose a novel self-supervised scheme to train an intrinsic-geometry-aware point feature extractor, which is illustrated in Fig.~\ref{fig:trainpipeline}. 

In particular, we assume to be given a set of human meshes as training data, and emphasize that no mesh input is used during inference. 
During training, we consume a pair of meshes $\s, \st$, and generate a subset of point clouds $\st_p$ from $\st$. 
We use ray-casting for partial shape generation, which is detailed in Sec.~\ref{sec:exp}. 

\noindent\paragraph{Pre-computing full-to-full maps: }As mentioned in Sec.~\ref{sec:intro}, our pipeline allows for both with and without correspondence annotations among the input training shapes. 
For the more non-trivial latter case, we simply train a Deep Functional Maps~\cite{dual} with the input meshes, and leverage its output (functional) maps in the following procedure. 

\noindent\paragraph{Backbone: } For extracting feature from (potentially partial) point clouds, we utilize the modified DGCNN proposed in~\cite{nie}, which is lightweight and robust with respect to point sampling density and distribution. In the following, we denote it by $\F$. 

\noindent\paragraph{Estimating Correspondence: } By passing $\s, \st_p$ through $\mathbf{F}$, we obtained the features $\F(\s), \F(\st_p)$, the point-wise correspondence can then be computed by searching for the nearest point in $\F(\s)$ of each query point in $\F(\st_p)$. 
However, for the sake of being differentiable, we adopt the following soft-map computation:
 \begin{equation}\label{eqn:pi}
\tilde{\Pi}_{\st_p\s}(i, j) = \frac{\exp(-{\alpha\delta_{ij}})}{\sum_{j'} \exp(-\alpha\delta_{ij'})}, \forall i\in [|\src|], j\in [|\st_p|], 
\end{equation}
where $\delta_{ij} = \|\F(\s)(i, :) -\F(\st_p)(j, :)\|$ and $\alpha$ is the temperature parameter, which is increased during training~\cite{dual}.


\noindent\paragraph{Training Loss: } We follow Eqn.~(\ref{eqn:steigen}) to compute the spatially truncated spectral embedding, $\Phi_{\st_p}$, for $\st_p$. 
As outlined above, we have access to the functional map $C_{\s\st}$, either given or estimated. 
According to Prop.~\ref{prop}, $C_{\s\st}$ is supposed to align $\Phi_{\st_p}$ and $\Phi_\s$, regardless of how $\st_p$ is generated. 
While it seems natural to formulate the following loss:
\[L(\F) = \arg\min_{\F} \Vert \Phi_{\st_p} C_{\s\st} - \tilde{\Pi}_{\st_p\s}\Phi_\s\Vert_{\mbox{Fro}}^2, \]
we notice that optimizing the above loss requires to compute Frobenius norm of a matrix of varying dimension, which depends on the number of points in $\st_p$. 
To alleviate such implementation difficulty, we propose the following two-step loss formulation. 
First we optimize $C_{\mbox{opt}}$ according to the following loss
\begin{equation}\label{eqn:l1}
    \begin{aligned}
        C_{\mbox{opt}}  =\arg\min_C &\left\|\Phi_{\st_p} C-\Pi_{\st_p\s} \Phi_{\s}\right\|_F^2\\
        &+\lambda\left\|\Delta_{\st} C-C \Delta_{\s}\right\|_F^2.
    \end{aligned}
\end{equation} 
In particular, since we estimate $\tilde{\Pi}_{\st_p\s}$ from learned features, in order to stabilize the noise during training, we adopt the regularization of commutativity with Laplacian (as the second term of the LHS of Eqn.(~\ref{eqn:l1})). 
The above optimization is linear in $C$, therefore admits a closed-form solution. 
Finally, the supervision from $C_{\s\st}$ is imposed by
\begin{equation}\label{eqn:l2}
    L(\F) = \Vert C_{\s\st} - C_{\mbox{opt}}\Vert_{\mbox{Fro}}^2.
\end{equation}

\noindent\textbf{Remark on the loss design.}
We note that our two-step loss formulation (Eqns.~\ref{eqn:l1}--\ref{eqn:l2}) offers several advantages over alternative supervision strategies.
First, directly comparing permutation matrices (\emph{e.g.,} $\Vert \tilde{\Pi}_{\st_p\s} - \Pi_{\st_p\s}^{*}\Vert$) requires access to ground-truth point-wise correspondences, which are unavailable in our self-supervised setting.
Second, comparing geodesic distance matrices would incur $O(n^2)$ computational cost per shape and is non-differentiable in general.
In contrast, our loss operates in the compact $k\times k$ spectral domain, admits a closed-form solution (detailed in the Supplementary Material), and naturally encodes structural priors such as bijectivity and Laplacian commutativity through the regularization in Eqn.~(\ref{eqn:l1}).

\subsubsection{Partial Shape Registration}\label{sec:reg}
In this section, for the sake of self-contentment, we entail the registration pipeline we adopt from~\cite{dfr}. 
During registration, the source template shape can be denoted by $\mathcal{S}^{k} = \{ \mathcal{V}^{k}, \mathcal{E} \}$,$ \mathcal{V}^{k}=\left\{ v_i^k \mid i=1,2, \ldots, N\right\}$, where $k$ indicates the iteration index and $v_i^k$ is the position of the $i-$th vertex at iteration $k$. The target partial point cloud is denoted by $\st_p =\left\{ u_j \mid j=1, \ldots, M\right\}$. 

Following ~\cite{dfr}, we first construct the deformation graph on template:$\mathcal{DG} = \{ 
\Theta\in \mathbb{R}^{H \times 3}, \Delta\in \mathbb{R}^{H \times 3}\, H = [N/2] \} $.  For a given $\textbf{X}^{k}$, we can compute displaced vertices via: 
 \begin{equation}\label{eqn:dg}
     \mathcal{V}^{k+1} = \mathcal{DG}(\textbf{X}^{k},\mathcal{V}^{k}).
 \end{equation}

Then we iteratively optimize the $\mathcal{DG}$ through a cost function. The energy terms in our registration pipeline are as follows:

\noindent\textbf{Correspondence Term} measures the distance of filtered correspondences between $\src^k$ and $\st_p$, given as:
\begin{equation}
      E_{\mbox{corr}} = \frac{1}{|\mathcal{C}^k|} \sum_{(v_i^k,u_j) \in \mathcal{C}^k }\left\| v^{k}_{i} - u_{j}\right\|_2^2, 
\end{equation}
where $\mathcal{C}^k $ is the filtered set of correspondences by a correspondence filtering module proposed in~\cite{dfr}

\noindent\textbf{Chamfer Distance Term} are to measure the extrinsic distance between $\src^k$ and $\st_p$. 
However, in our case, the target point cloud in general corresponds to a true subset of the source template, which naturally leads to the following modified version of the Chamfer Distance:

\begin{equation}
    E_{\mbox{cd}}=\frac{1}{M} \sum_{ j \in [M]} \min_{ i \in [N] } \left\| v^{k}_{i} - u_{j}     \right\|_2^2.
\end{equation}

\noindent\textbf{As-Rigid-As-Possible Term} reflects the deviation of estimated local surface deformations from rigid transformations. 
We follow \cite{dfr,guo2021human, levi2014smooth} and define it as:
\begin{equation}
    E_{\mbox{arap}} =\sum_{h \in [H]} \sum_{l \in \psi(h)}(\left\|d_{h, l}( \textbf{X})\right\|_2^2  + \beta \left\|(R\left( \Theta_h\right)
    -R\left( \Theta_l \right)\right\|_2^2 )
\end{equation} 
\begin{align}
\begin{aligned}
     d_{h, l}( \textbf{X}) 
     &=d_{h, l}( \Theta,\Delta)\\
     &= R\left( \Theta _h\right)\left( g _l- g _h\right)+ \Delta _k+ g _k-\left( g _l+ \Delta _l\right).   
\end{aligned}
\end{align}
Here, $g \in R^{H\times 3}$ are the original positions of the nodes in the deformation graph $\mathcal{DG}$, and $\psi(h)$ is the 1-ring neighborhood of the $h-$th deformation node. $R(\cdot)$ is Rodrigues' rotation formula that computes a rotation matrix from an axis-angle representation and $\beta$ is the weight of the smooth rotation regularization term.

 \noindent\textbf{Total cost function:} The total cost function follows the same formulation as in DFR (Eqn.~\ref{eqn:total}), with $E_{\mbox{cd}}$ replaced by the unidirectional variant defined above.
We refer readers to Alg.~1 of~\cite{dfr} for more details of the algorithm for minimizing $E_{\mbox{total}}$.

\section{Experimental Results}
We undertake a series of experiments employing DFR and its variant, Partial-DFR, to investigate the registration of full shape and partial shapes, 
respectively.
\label{sec:exp}
\subsection{Experimental Results of DFR}
In this subsection, we present the results of DFR from our previous work~\cite{dfr}, highlighting its performance under the \emph{full-2-full} setting.

\noindent\textbf{Datasets:} We evaluate our method and several state-of-the-art techniques for estimating correspondences between deformable shapes on an array of benchmarks as follows: \textbf{FAUST\_r: }The remeshed version of FAUST dataset~\cite{bogo2014}, which consists of 100 human shapes (10 individuals performing the same 10 actions). We split the first 80 as training shapes and the rest as testing shapes; \textbf{SCAPE\_r: }The remeshed version of SCAPE dataset~\cite{scape}, which consists 71 human shapes (same individual in varying poses). We split the first 51 as training shapes and the rest as testing shapes; \textbf{SHREC19\_r: } The remeshed version of SHREC19 dataset~\cite{SHREC19}, which consists of 44 shapes of different identities and poses. We use it solely in test, and follow the test pair list provided by~\cite{SHREC19}; \textbf{SHREC07-H: } A subset of SHREC07 dataset~\cite{shrec07}, which consists of 20 heterogeneous human shapes of the varying number of vertices. We use it solely in test, and use the accompanied sparse landmark annotations to quantify all the pairwise maps among them; \textbf{DT4D-H: } A dataset proposed in~\cite{magnet2022smooth}, which consists of 10 categories of heterogeneous humanoid shapes. We use it solely in testing, and evaluating the inter-class maps split in~\cite{li2022attentivefmaps}; \textbf{TOPKIDS: } This is a challenging dataset~\cite{lahner2016shrec} consisting of 26 shapes of a kid in different poses, which manifest significant topological perturbations in meshes. 

\noindent\textbf{Baselines: } We compare our method with an array of competitive baselines, including axiomatic shape registration methods: Smooth Shells~\cite{smoothshells}, NDP~\cite{li2022non}, AMM~\cite{AMM}; learning-based registration methods: 3D-CODED~\cite{groueix20183d}, Deep Shells~\cite{deepshells}, TransMatch~\cite{trappolini2021shape}, SyNoRiM~\cite{huang_multiway_2022}; deep functional maps frameworks: SURFMNet~\cite{unsuperise_fmap}, WSupFMNet~\cite{sharma20}, GeomFMaps~\cite{donati20}, DiffFMaps~\cite{lie}, NIE~\cite{nie}, ConsistFMaps~\cite{cao2022}, SSMSM~\cite{cao2023}. According to input requirements, we put those relying on pure mesh input on the top, and the rest, suitable for point clouds, in the bottom of both tables.

\noindent\textbf{Train/Test Cases:} Throughout this section, we consider a highly challenging scenario: for each learning-based method, we train two models respectively with FAUST\_r and SCAPE\_r dataset, and then we run tests in a range of test cases including FAUST\_r, SCAPE\_r, SHREC19\_r, SHREC07-H, DT4D-H and TOPKIDS. The test pairs of each case have been described above. In the following, $A/B$ means we train on dataset $A$ and test on dataset $B$. 

\noindent\textbf{Datasets Alignment: }There exist extrinsically aligned versions for all the involved datasets but SHREC07-H. We equally feed in aligned datasets to all the baselines when available. On the other hand, the original version of SHREC07-H manifests significant variability of orientations. For the sake of fairness and simplicity, we apply our orientation regressor to it and provide baselines for our automatically aligned data. Note that, all the aligned datasets, as well as the synthetic dataset on which we train our orientation regressor, roughly share the same canonical orientation, which is defined by the SMPL~\cite{SMPL} model. 
Finally, we always apply automatic alignment before inputting shapes into our pipeline, whatever the original orientation they are. 

\noindent\textbf{Choice of Source Shape: }As mentioned at the beginning of Sec.~\ref{sec:method}, we compute correspondences between two raw input point clouds by associating them via a common source mesh $\src$. In Tab.~\ref{table:iso} and Tab.~\ref{table:noniso}, we indicate our choice of source mesh by Ours-\emph{name}, where \emph{name} indicates the origin of the source mesh. 
For simplicity, we fix the source mesh from each dataset and visualize them in the appendix.
On the other hand, when implementing axiomatic shape registration methods, we only consider deforming the same template we pick from FAUST\_r (resp. SCAPE\_r) to every test point cloud in the test set of FAUST\_r (resp. SCAPE\_r), and establish correspondence by map composition, in the same manner as ours. 

\noindent\textbf{Metric: }Though we primarily focus on matching point clouds, we adopt the commonly used geodesic error normalized by the square root of the total area of the mesh, to evaluate all methods, either for mesh or point cloud. 

\noindent\textbf{Hyper-Parameters: } We remark that all the hyper-parameters are fixed for \emph{all} experiments in this section. In particular, we settle them by performing a grid search with respect to the weights used in the final optimization to seek for the combination that leads to the best registration results (quantitatively in terms of Chamfer distance and qualitatively by visual inspection) on a few training shapes.
Specifically, for \textbf{full registration} (DFR), we use $\lambda_{\mbox{corr}}=1, \lambda_{\mbox{cd}}=0.01, \lambda_{\mbox{arap}}=20$ in Stage-I and $\lambda_{\mbox{corr}}=0.01, \lambda_{\mbox{cd}}=1, \lambda_{\mbox{arap}}=1$ in Stage-II.
For \textbf{partial registration} (Partial-DFR), we use $\lambda_{\mbox{corr}}=0.01, \lambda_{\mbox{cd}}=1, \lambda_{\mbox{arap}}=20$ in Stage-I and $\lambda_{\mbox{corr}}=0.01, \lambda_{\mbox{cd}}=1, \lambda_{\mbox{arap}}=1$ in Stage-II.
The key difference is in Stage-I: in partial registration, $\lambda_{\mbox{cd}}$ is significantly larger ($1$ vs. $0.01$) and $\lambda_{\mbox{corr}}$ is smaller ($0.01$ vs. $1$). This is because the switch from bidirectional to unidirectional Chamfer distance reduces the magnitude of the CD term (only target-to-source distances are summed), which is compensated by increasing $\lambda_{\mbox{cd}}$. Meanwhile, correspondences on partial shapes are inherently noisier, motivating a smaller $\lambda_{\mbox{corr}}$ to avoid propagating erroneous correspondences.
We provide more detailed discussion and ablation of the choice of hyper-parameters in the appendix.

\subsubsection{Experimental Results}
Subsequently, we conduct the experiments of full shape registration in terms of Near-isometric Benchmarks, Non-isometric Benchmarks and Topologically Perturbed Benchmark.
\noindent\paragraph{Near-isometric Benchmarks}\label{sec:iso}
As shown in Fig.~\ref{fig:teaser}, in the presence of large deformation between the template and the target, NDP\cite{li2022non} and AMM\cite{AMM} fail completely while our method delivers high-quality deformation. 

Moreover, as illustrated in Tab.~\ref{table:iso}, our method achieves the best performance in 5 out of 6 test cases. Remarkably, in the two most challenging tasks, FAUST\_r/SHREC19\_r and FAUST\_r/SHREC19\_r, our method indeed outperforms \emph{all} of the baselines, including the state-of-the-art methods that take meshes as input. Regarding point-based methods, SSMSM~\cite{cao2023} performs well in the standard case and outperforms ours in FASUT\_r/FAUST\_r, but generalizes poorly to unseen shapes. Another important observation is that our method manifests robustness with respect to the choice of template shapes. In fact, the above observations remain true no matter which template we select. 

\begin{table}[t!]
\caption{Mean geodesic errors (×100) on FAUST\_r (F\_r), SCAPE\_r (S\_r) and SHREC19\_r (S19\_r). The \textbf{best} is highlighted. }\label{table:iso}
\centering
\scriptsize
\setlength{\tabcolsep}{1.6mm}
\begin{tabular}{clccclccc}
\hline
\rowcolor[HTML]{FFFFFF} 
\cellcolor[HTML]{FFFFFF}                         & \multicolumn{1}{c}{\cellcolor[HTML]{FFFFFF}Train}     & \multicolumn{3}{c}{\cellcolor[HTML]{FFFFFF}F\_r}                                                   &                                                       & \multicolumn{3}{c}{\cellcolor[HTML]{FFFFFF}S\_r}                                                   \\
\rowcolor[HTML]{FFFFFF} 
\multirow{-2}{*}{\cellcolor[HTML]{FFFFFF}Method} & \multicolumn{1}{c}{\cellcolor[HTML]{FFFFFF}Test}      & F\_r                     & S\_r                     & S19\_r                               & \multicolumn{1}{c}{\cellcolor[HTML]{FFFFFF}}          & S\_r                    & F\_r                     & S19\_r                               \\ \hline
\rowcolor[HTML]{FFFFFF} 
\rowcolor[HTML]{FFFFFF} 
Smooth Shells~\cite{smoothshells}                                    &                                                       & 2.5                          & \textbackslash{}             & \textbackslash{}                         &                                                       & 4.7                          & \textbackslash{}             & \textbackslash{}                         \\
\rowcolor[HTML]{FFFFFF} 
SURFMNet(U)~\cite{unsuperise_fmap}                                      & \multicolumn{1}{c}{\cellcolor[HTML]{FFFFFF}}          & 15.0                         & 32.0                         & \textbackslash{}                         &                                                       & 12.0                         & 32.0                         & \textbackslash{}                         \\
\rowcolor[HTML]{FFFFFF} 
\rowcolor[HTML]{FFFFFF} 
NeuroMorph(U)~\cite{eisenberger2021neuromorph}                                    & \multicolumn{1}{c}{\cellcolor[HTML]{FFFFFF}}          & 8.5                          & 29.0                         & \textbackslash{}                         &                                                       & 30.0                         & 18.0                         & \textbackslash{}                         \\
\rowcolor[HTML]{FFFFFF} 
\rowcolor[HTML]{FFFFFF} 
WSupFMNet(W)~\cite{sharma20}                                     & \multicolumn{1}{c}{\cellcolor[HTML]{FFFFFF}}          & 3.3                          & 12.0                         & \textbf{\textbackslash{}}                &                                                       & 7.3                          & 6.2                          & \textbf{\textbackslash{}}                \\
\rowcolor[HTML]{FFFFFF} 
GeomFMaps(S)~\cite{donati20}                                     & mesh          & 3.1                          & 11.0                         & 9.6                                      &                                                       & 4.4                          & 6.0                          & 11.4                                     \\
\rowcolor[HTML]{FFFFFF} 
Deep Shells(W)~\cite{deepshells}                                  & \multicolumn{1}{c}{\cellcolor[HTML]{FFFFFF}}          & 1.7                          & 5.4                          & 26.6                                     &                                                       & 2.5                          & 2.7                          & 21.4                                     \\
\rowcolor[HTML]{FFFFFF} 
ConsistFMaps(U)~\cite{cao2022}                                         & \multicolumn{1}{c}{\cellcolor[HTML]{FFFFFF}\textbf{}} & \textbf{1.5}                 & 7.3                          & 20.9                                     &                                                       & \textbf{2.0}                 & 8.6                          & 28.7                                     \\
\rowcolor[HTML]{FFFFFF} 
AttentiveFMaps(U)~\cite{li2022attentivefmaps}                                &                                                       & 1.9                          & \textbf{2.6}                 & \textbf{6.2}                             &                                                       & 2.2                          & \textbf{2.2}                 & \textbf{9.3}                             \\ \hline
\rowcolor[HTML]{FFFFFF} 
NDP~\cite{li2022non}                                              &                                                       & 20.4                         & \textbackslash{}             & \textbackslash{}                         &                                                       & 16.2                         & \textbackslash{}             & \textbackslash{}                         \\
\rowcolor[HTML]{FFFFFF} 
AMM~\cite{AMM}                                              &                                                       & 14.2                         & \textbackslash{}             & \textbackslash{}                         &                                                       & 13.1                         & \textbackslash{}             & \textbackslash{}                         \\
\cellcolor[HTML]{FFFFFF}CorrNet3D(U)~\cite{zeng2021corrnet3d}             & \cellcolor[HTML]{FFFFFF}                              & \cellcolor[HTML]{FFFFFF}63.0 & \cellcolor[HTML]{FFFFFF}58.0 & \cellcolor[HTML]{FFFFFF}\textbackslash{} &                                                       & \cellcolor[HTML]{FFFFFF}58.0 & \cellcolor[HTML]{FFFFFF}63.0 & \cellcolor[HTML]{FFFFFF}\textbackslash{} \\
\rowcolor[HTML]{FFFFFF} 
3D-CODED(S)~\cite{groueix20183d}                                      & \multicolumn{1}{c}{\cellcolor[HTML]{FFFFFF}}          & 2.5                          & 31.0                         & \textbackslash{}                         &                                                       & 31.0                         & 33.0                         & \textbackslash{}                         \\
\rowcolor[HTML]{FFFFFF} 
SyNoRiM(S)~\cite{huang_multiway_2022}                                       &                                                       & 7.9                          & 21.9                         & \textbackslash{}                         &                                                       & 9.5                          & 24.6                         & \textbackslash{}                         \\
\cellcolor[HTML]{FFFFFF}TransMatch(S)~\cite{trappolini2021shape}            & pcd   & 2.7                          & \cellcolor[HTML]{FFFFFF}33.6 & \cellcolor[HTML]{FFFFFF}21.0             & \multicolumn{1}{c}{\cellcolor[HTML]{FFFFFF}\textbf{}} & \cellcolor[HTML]{FFFFFF}18.6 & \cellcolor[HTML]{FFFFFF}18.3 & \cellcolor[HTML]{FFFFFF}38.8             \\

\rowcolor[HTML]{FFFFFF} 
DPC(S)~\cite{lang2021dpc}                                           &                                                       & 11.1                          & 17.5                         & 31.0                                     &                                                       & 17.3                         & 11.2                         & 28.7                                     \\

\rowcolor[HTML]{FFFFFF} 
DiffFMaps(S)~\cite{lie}                                           &                                                       & 3.6                          & 19.0                         & 16.4                                     &                                                       & 12.0                         & 12.0                         & 17.6                                     \\
\rowcolor[HTML]{FFFFFF} 
NIE(W)~\cite{nie}                                           &                                                       & 5.5                          & 15.0                         & 15.1                                     &                                                       & 11.0                         & 8.7                          & 15.6                                     \\
\rowcolor[HTML]{FFFFFF} 
SSMSM(W)~\cite{cao2023}                                         &                                                       & \textbf{2.4}                 & 11.0                         & 9.0                                      &                                                       & 4.1                          & 8.5                          & 7.3                                      \\
\rowcolor[HTML]{FFFFFF} 
\rowcolor[HTML]{E7E6E6} 
Ours-SCAPE                                       &                                                       & 3.4                          & \textbf{5.1}                 & \textbf{5.4}                             &                                                       & \textbf{2.6}                 & \textbf{4.0}                 & 5.1                                      \\
\rowcolor[HTML]{E7E6E6} 
Ours-FAUST                                       &                                                       & 3.0                          & 6.3                          & 5.9                                      &                                                       & 2.9                          & 4.0                          & \textbf{4.8}                             \\
 \hline
\end{tabular}
\vspace{-0.8em}
\end{table}
\begin{table}[t!]
\caption{Mean geodesic errors (×100) on SHREC’07-H and DT4D-H. The \textbf{best} is highlighted. }\label{table:noniso}
\centering
\scriptsize
\begin{tabular}{ccccccc}
\hline
\rowcolor[HTML]{FFFFFF} 
\cellcolor[HTML]{FFFFFF}                         & Train                    & \multicolumn{2}{c}{\cellcolor[HTML]{FFFFFF}FAUST\_r}        &                                              & \multicolumn{2}{c}{\cellcolor[HTML]{FFFFFF}SCAPE\_r}                                                        \\ 
\rowcolor[HTML]{FFFFFF} 
\multirow{-2}{*}{\cellcolor[HTML]{FFFFFF}Method} & Test                     & SHREC07                   & DT4D                      & \multicolumn{1}{l}{\cellcolor[HTML]{FFFFFF}} & \multicolumn{1}{l}{\cellcolor[HTML]{FFFFFF}SHREC07} & \multicolumn{1}{l}{\cellcolor[HTML]{FFFFFF}DT4D} \\ \hline
\rowcolor[HTML]{FFFFFF} 
GeomFMaps~\cite{donati20}                                        &                          & 30.5                         & 38.5                         &                                              & 28.9                                                   & 28.6                                               \\
\rowcolor[HTML]{FFFFFF} 
Deep Shells~\cite{deepshells}                                     &                       & 30.6                         & 35.9                         &                                              & 31.3                                                   & 25.8                                               \\
\rowcolor[HTML]{FFFFFF} 
ConsistFMaps~\cite{cao2022}                                            &   mesh                       & 36.2                         & 33.5                         &                                              & 37.3                                                   & 38.6                                               \\
\rowcolor[HTML]{FFFFFF} 
AttentiveFMaps~\cite{li2022attentivefmaps}                                   &                          & \textbf{16.4}                & \textbf{11.0}                & \textbf{}                                    & \textbf{21.1}                                          & \textbf{21.4}                                      \\ \hline
\cellcolor[HTML]{FFFFFF}TransMatch~\cite{trappolini2021shape}               & \cellcolor[HTML]{FFFFFF} & \cellcolor[HTML]{FFFFFF}25.3 & \cellcolor[HTML]{FFFFFF}26.7 & \cellcolor[HTML]{FFFFFF}                     & 31.2                                                   & \cellcolor[HTML]{FFFFFF}25.3                       \\
\rowcolor[HTML]{FFFFFF} 
DiffFMaps~\cite{lie}                                              &                          & 16.8                         & 18.5                         &                                              & 15.4                                                   & 15.9                                               \\
\rowcolor[HTML]{FFFFFF} 
NIE~\cite{nie}                                              &                          & 15.3                         & 13.3                         &                                              & 13.4                                                   & 12.1                                               \\
\rowcolor[HTML]{FFFFFF} 
SSMSM~\cite{cao2023}                                            &  pcd                        & 42.2                         & 11.8                         &                                              & 37.7                                                   & 8.0                                                \\
\rowcolor[HTML]{E7E6E6} 
Ours-SCAPE                                       &                          & 9.3                          & 9.8                          &                                              & \textbf{5.9}                                           & 5.7                                                \\
\rowcolor[HTML]{E7E6E6} 
Ours-FAUST                                       & \textbf{}                & \textbf{8.5}                 & \textbackslash{}             &                                              & 6.1                                                    & \textbackslash{}                                   \\
\rowcolor[HTML]{E7E6E6} 
Ours-CRYPTO                                      & \textbf{}                & \textbackslash{}             & \textbf{6.9}                 & \multicolumn{1}{l}{\cellcolor[HTML]{E7E6E6}} & \textbackslash{}                                       & \textbf{5.7}                                       \\ \hline
\end{tabular}
\vspace{-2em}
\end{table}
\noindent\paragraph{Non-isometric Benchmarks}\label{sec:non-iso}
We stress test our method on challenging non-isometric datasets including SHREC07-H and DT4D-H. We emphasize that these test shapes are unseen during training. SHREC07-H contains 20 heterogeneous human shapes, whose number of vertices ranges from $3000$ to $15000$. Moreover, there exists some topological noise (\emph{e.g.}, the hands of the rightmost shape in Fig.~\ref{fig:teaser} is attached to the thigh in mesh representation). As shown in Fig.~\ref{fig:teaser}, SSMSM~\cite{cao2023} barely delivers reasonable results, which might be due to the sensitivity of graph Laplacian construction on point clouds. Topological noise, on the other hand, degrades mesh-based methods like Deep Shells~\cite{deepshells}. Meanwhile, as shown in Tab.~\ref{table:noniso}, our method achieves a performance improvement of over 40\% compared to the previous SOTA approaches ($\mathbf{8.5}$ vs. 15.3; $\mathbf{5.9}$ vs. 13.4), which again confirms the robustness of our approach. Regarding DT4D-H, We follow the test setting of  AttentiveFMaps~\cite{li2022attentivefmaps}, and only consider the more challenging inter-class mapping. Interestingly, our method outperforms the state-of-the-art point cloud-based approach by approximately 30\% and exceeds the performance of mesh-based state-of-the-art methods by over 70\%. 
Furthermore, even in the worst case, training on FAUST\_r and using SCAPE template, our error is still the lowest compared to external baselines. 

\begin{table*}[h!]
\caption{Mean geodesic errors (×100) on TOPKIDS which trained on FAUST\_r and SCAPE\_r. The \textbf{best} is highlighted. }\label{table:topkids}
\centering
\scriptsize
\setlength{\tabcolsep}{2mm}
\resizebox{\textwidth}{5mm}{
\begin{tabular}{
>{\columncolor[HTML]{FFFFFF}}c 
>{\columncolor[HTML]{FFFFFF}}c 
>{\columncolor[HTML]{FFFFFF}}c 
>{\columncolor[HTML]{FFFFFF}}c 
>{\columncolor[HTML]{FFFFFF}}c |
>{\columncolor[HTML]{FFFFFF}}c 
>{\columncolor[HTML]{FFFFFF}}c 
>{\columncolor[HTML]{FFFFFF}}c 
>{\columncolor[HTML]{E7E6E6}}c }
\hline
                                & GeomFMaps~\cite{donati20}  & Deep Shells~\cite{deepshells} & ConsistFMaps~\cite{cao2022} & AttentiveFMaps~\cite{li2022attentivefmaps} & DiffFMaps~\cite{lie} & NIE~\cite{nie}  & SSMSM~\cite{cao2023} & Ours         \\ \hline
FAUST\_r\textbackslash{}TOPKIDS & 26.2               & \textbf{14.7}        & 35.9                  & 31.7                    & 20.5      & 18.9 & 14.2  & \textbf{8.9} \\
SCAPE\_r\textbackslash{}TOPKIDS & 21.7               & \textbf{15.3}        & 33.1                  & 39.4                    & 18.0      & 16.2 & 12.3  & \textbf{7.1} \\ \hline
\end{tabular}}
\vspace{-3em}
\end{table*}
\noindent\paragraph{Topologically Perturbed Benchmark}
The shapes in TOPKIDS present various adhesions -- hand/arm adhered to abdomen/thigh. Given the ground-truth maps from each topologically perturbed shape to the reference shape, we evaluate all 25 maps (excluding the trivial one from reference to itself) with models trained on FAUST\_r and SCAPE\_r respectively. Since the orientation of the TOPKIDS dataset is not originally agreeable with the aligned version of the two training sets, we align the input with our orientation regressor and feed them into the baselines depending on extrinsic information~\cite{deepshells, lie, nie, cao2023}. Regarding the template shape, we adopt the reference shape as our source mesh in registration. We report the quantitative comparisons in Tab.~\ref{table:topkids}. It is obvious that topological noise poses great challenges for methods based on pure intrinsic information~\cite{donati20, cao2022, li2022attentivefmaps}, while the counterparts perform relatively well. Especially, among the point-based methods, our method outperforms the latest SOTA method by a large margin, namely, over 40\% relative error reduction  ($\mathbf{7.1}$ vs. $12.3$). 
We refer readers to the appendix for a qualitative comparison, which agrees with the quantitative results above. 

\subsubsection{Ablation Study}\label{sec:abl}
\begin{table}[!t]
\caption{Mean geodesic errors (×100) on different ablated settings,
the models are all trained on SCAPE\_r and test on SHREC'07.}\label{table:abl}
\centering
\scriptsize
\setlength{\tabcolsep}{0.7mm}
\begin{tabular}{
>{\columncolor[HTML]{FFFFFF}}c 
>{\columncolor[HTML]{FFFFFF}}c
>{\columncolor[HTML]{FFFFFF}}c 
>{\columncolor[HTML]{FFFFFF}}c 
>{\columncolor[HTML]{FFFFFF}}c 
>{\columncolor[HTML]{FFFFFF}}c 
>{\columncolor[HTML]{FFFFFF}}c }
\hline
w/o Registration & w/o Stage I & w/o Stage II & w/o updating corre. & w/o cons. filter  & Full \\ \hline
 11.5 & 10.6 & 10.1           & 8.1                 & 7.2                                                 & 5.9  \\ \hline
\end{tabular}
\vspace{-1em}
\end{table}

We report ablation studies in Tab.~\ref{table:abl}, in which we verify the effectiveness of each core design formulated in Sec.~\ref{sec:method}. Throughout this part, we train on SCAPE\_r and test on SHREC07-H. In the registration stage, we use the SCAPE\_r template as the source mesh and deform it to each point cloud from SHREC07-H. It is evident that each ablated module contributes to the final performance. In particular, both stages of registration play an equally important role in our framework.

 Finally, to validate the superiority of our registration scheme, we report in Tab.~\ref{table:reg} the average errors of the initial maps computed by our point-based DFM, and that of output maps of Ours, NDP~\cite{li2022non}, AMM~\cite{AMM}, which are all based on the former. It is evident that, across three different test sets, our method consistently improves the initial maps, while NDP and AMM can even lead to deteriorated maps than the initial input.

\begin{table}[!t]
    \caption{Mean geodesic errors (×100) of Ours, NDP, AMM based on the same initial maps.}\label{table:reg}
    \centering
    \scriptsize
    \begin{tabular}{cccc}
    \hline
    \rowcolor[HTML]{FFFFFF} 
    \multicolumn{1}{l}{\cellcolor[HTML]{FFFFFF}} & SCAPE\_r         & SHREC19\_r    & SHREC07-H     \\ \hline
    \rowcolor[HTML]{FFFFFF} 
    Ini.                                    & 5.5         & 8.1         & 11.5         \\
    \rowcolor[HTML]{FFFFFF} 
    NDP                                         & 5.4        &11.4       & 8.9     \\
    \rowcolor[HTML]{FFFFFF} 
    AMM                                        & 11.4       & 10.7        & 8.8        \\
    \rowcolor[HTML]{E7E6E6} 
    Ours                                         & \textbf{2.6} & \textbf{5.1} & \textbf{5.9} \\ \hline
    \end{tabular}
    \vspace{-0.8em}
\end{table}

\subsection{Experimental Results of Partial-DFR}
\subsubsection{Model Training}
In this section, we train two versions of point feature extractor using different datasets and correspondence annotations. 
To generate partial shapes for training and test, we render 12 partial-view point clouds for each full shape from the training set, oriented in the directions from $12$ vertices of a regular icosahedron to the origin, and obtain the point clouds via ray-casting. 
Unlike previous partial-view datasets such as those in ~\cite{huang20, cao2023}, which only render shapes around the z-axis, our dataset features a diverse range of connectedness, partiality patterns, and sizes.

\noindent\paragraph{Ours-S\&F} is trained on \textbf{S$\&$F}, an assembled dataset consisting of the remeshed version of FAUST dataset~\cite{bogo2014} and SCAPE dataset~\cite{scape}. We merge the training and testing shapes of the two datasets which in total consists of $171$ human shapes. We split the first $80$ of FAUST\_r and $51$ of SCAPE\_r as training shapes and the rest as test shapes. 
Regarding full shape correspondence annotation, we utilize  full-to-full maps pre-computed by an unsupervised DFM~\cite{dual} for supervision, eliminating the need for any ground-truth correspondence. 

\noindent\paragraph{Ours-M\&P} is trained on $340$ shapes sampled from the MoSh and PosePrior sequences, which are included in the AMASS~\cite{amass} dataset. 
As the shapes from AMASS dataset are all in identity maps with each other, we directly leverage them as the supervision signal. 

\noindent\paragraph{Integration with Neural Enhanced Shape Registration: }Throughout the experimental results, unless otherwise specified (Sec.~\ref{sec:abl}), we always report the matching score of the full pipeline combining our learned point-feature extractor with the registration component proposed in~\cite{dfr}. 

\noindent\paragraph{Baselines:} We compare our method with an array of competitive baselines, including axiomatic shape registration methods:  NDP~\cite{li2022non}, AMM~\cite{yao2023fast}; learning-based registration methods: DFR~\cite{dfr},HCLV2S~\cite{huang20}; deep functional maps frameworks: DPFM~\cite{attaiki2021dpfm}, GeomFMaps~\cite{donati20}, DiffFMaps~\cite{lie}, NIE~\cite{nie} SSMSM~\cite{cao2023}. 


\subsubsection{Partial Shape Matching without Correspondence Label}

\begin{table}[!t]
    \caption{Mean geodesic errors (×100) on SCAPE-FAUST partial dataset from regular icosahedron views. The \textbf{best} is highlighted.}
    \centering
    \scriptsize
    \begin{tabular}{cccc}
        \toprule
        Method  & Geo. error(×100) \\
        \midrule
        DPFM~\cite{attaiki2021dpfm} unsup	&13.15	\\
        DPFM~\cite{attaiki2021dpfm} sup	&5.18	\\
        NIE~\cite{nie}   &11.17  \\
        DiffFmaps~\cite{lie}   &12.04  \\
        SSMSM~\cite{cao2023}   &8.12  \\
        HCLV2S~\cite{huang20}	&6.98	\\
        \rowcolor[HTML]{E7E6E6} 
        Ours	&\textbf{2.33}	\\
        \hline
    \end{tabular}\label{table:sf}
    \vspace{-0.8em}
\end{table}
In this section, we train and test all the baselines in the same way as we train \textbf{Ours-S\&F}, with an exception of HCLV2S~\cite{huang20}, as the latter requires one-to-one correspondences between a template and all training partial shapes, which is unsatisfied in the \textbf{S$\&$F} dataset.
Instead we evaluate with the official checkpoint from HCLV2S~\cite{huang20}, which is trained on $190$K shapes from SURREAL dataset. 

Tab.~\ref{table:sf} reports the quantitative results of our method and several recent baselines, where our method outperforms other methods by a large margin ($55.0\%$ error reduction compared to the second best, \emph{supervised} DPFM~\cite{attaiki2021dpfm}). 
It is noteworthy that the DPFM is designed for connected meshes and inherently underperforms when handling meshes with disconnected regions. During both training and testing, we consistently retain only the largest connected region of each partial shape. Despite the extra care, the supervised version of DPFM still performs significantly worse than our method, let alone the unsupervised version of DPFM. 
The results reveal that, despite of significantly larger training set and correspondence supervision, it still struggles to generalize to unseen diverse partial views effectively.

\subsubsection{Stress Tests on External Tasks}
In this part, we stress test our model on three external challenging tasks from~\cite{attaiki2021dpfm,cao2023, huang20} respectively. 
We emphasize that our model are trained without any knowledge of the task or the test data. 

\begin{table}[!t]
    \caption{Mean geodesic errors (×100) comparison with other methods on PFARM dataset. The \textbf{best} is highlighted.}
    \centering
    \scriptsize
    \begin{tabular}{cccc}
        \toprule
        Method  & Geo. error(×100) \\
        \midrule
        PFM(+ZO)~\cite{rodola2016partial}	&42.34	\\
        FSP(+ZO)~\cite{litany2017fully}	&53.15	\\
        DOC(+ZO)~\cite{cosmo2016matching}	&51.78	\\
        GeomFMaps(+ZO)~\cite{donati20}	&22.22	\\
        DPFM(+ZO)~\cite{attaiki2021dpfm}	&10.53	\\
        SSMSM~\cite{cao2023}	&12.32	\\
        \rowcolor[HTML]{E7E6E6} 
        Ours-S$\&$F & \textbf{7.35}	\\
        \rowcolor[HTML]{E7E6E6} 
        Ours-M$\&$P	& 8.30	\\
        \hline
    \end{tabular}\label{table:pfarm}
    \vspace{-0.8em}
\end{table}
\noindent\paragraph{PFARM} In the first task, we consider the PFARM benchmark proposed by DPFM~\cite{attaiki2021dpfm}, where the benchmark is used for testing generalizability of methods trained on \textbf{CUTS}~\cite{attaiki2021dpfm}. 
Tab.~\ref{table:pfarm} reports the test error of our two models, we highlight that our training shapes are all obtained from partial views evenly distributed in the sphere around the training shapes, therefore Cut partiality is alien to our models. 
Yet, either of our model achieves at least $30\%$ improvement over the baselines listed in~\cite{attaiki2021dpfm}. 
Moreover, all the baselines in Tab.~\ref{table:pfarm} utilize ZoomOut~\cite{zoomout} for post-processing, we refer readers to the Supplementary Material for the raw output performance. 

\begin{table}[!t]
    \caption{Mean geodesic errors (×100) on SURREAL partial view dataset from SSMSM. The \textbf{best} is highlighted.}
    \centering
    \scriptsize
    \begin{tabular}{cccc}
        \toprule
        Method  & Geo. error(×100) \\
        \midrule
        DPFM~\cite{attaiki2021dpfm} unsup	&12.0	\\
        DPFM~\cite{attaiki2021dpfm} sup	&7.8	\\
        SSMSM~\cite{cao2023}	&6.3	\\
        \rowcolor[HTML]{E7E6E6} 
        Ours-S$\&$F	&6.3	\\
        \rowcolor[HTML]{E7E6E6} 
        Ours-M$\&$P	&\textbf{4.9}	\\
        \hline
    \end{tabular}\label{table:ssmsm}
    \vspace{-0.8em}
\end{table}
\noindent\paragraph{Surreal-PV} In the second task, we test with the benchmark proposed in SSMSM~\cite{cao2023}, which consists of $5000$ partial view point clouds cropped from SURREAL dataset. 
In particular, the baselines in Tab.~\ref{table:ssmsm} are all trained on $4000$ shapes from it and test on the rest. 
In contrast, we \emph{directly} test our model with the $1000$ point clouds. 
Remarkably, the model trained on $171$ shapes, \textbf{Ours-S\&F} has already achieved on-par performance with~\cite{cao2023}. 
Furthermore, our model trained on the larger dataset surpass the second best method by $22\%$.
\begin{table}[t!]
\caption{Euclidean distance evaluation for partial-2-full correspondence task. We report average correspondence error, 5cm-recall and 10cm-recall on FAUST and SHREC'19 dataset from HCLV2S\cite{huang20} baseline. The \textbf{best} is highlighted.}\label{table:hcv2s}
\centering
\scriptsize
\setlength{\tabcolsep}{0.6mm}
\begin{tabular}{clccclccc}
\hline
\rowcolor[HTML]{FFFFFF} 
\cellcolor[HTML]{FFFFFF}                         & \multicolumn{1}{c}{\cellcolor[HTML]{FFFFFF}}     & \multicolumn{3}{c}{\cellcolor[HTML]{FFFFFF}FAUST}                                                   &                                                       & \multicolumn{3}{c}{\cellcolor[HTML]{FFFFFF}SHREC19}                                                   \\ 
\rowcolor[HTML]{FFFFFF} 
\multirow{-2}{*}{\cellcolor[HTML]{FFFFFF}Method} & \multicolumn{1}{c}{\cellcolor[HTML]{FFFFFF}}      & AE(cm)     & 5cm-recall                 & 10cm-recall                             & \multicolumn{1}{c}{\cellcolor[HTML]{FFFFFF}}         & AE(cm)     & 5cm-recall                 & 10cm-recall                               \\ \hline
\rowcolor[HTML]{FFFFFF} 
\text{DHBC~\cite{wei2016dense}}                                   &                                                       & 10.91                          & 0.503           & 0.772                         &                                                       & 17.24                         & 0.401           & 0.646                    \\
\rowcolor[HTML]{FFFFFF} 
\text{SMPL}~\cite{SMPL}                             & \multicolumn{1}{c}{\cellcolor[HTML]{FFFFFF}}          & 1.98                   & 0.932                     & 0.973                     &                                                       & 5.48                         & 0.751                         & 0.897
\\
\rowcolor[HTML]{FFFFFF} 
\text{HCLV2S~\cite{huang20}}                           & \multicolumn{1}{c}{\cellcolor[HTML]{FFFFFF}}          & 1.90                         & 0.953                         & \textbf{0.998}                         &                                                       & 4.81                       & 0.810                        & \textbf{0.970}   \\
\rowcolor[HTML]{E7E6E6} 
Ours-M$\&$P                        & \multicolumn{1}{c}{\cellcolor[HTML]{E7E6E6} }          & \textbf{1.46}                          & \textbf{0.968}                         & 0.996                        &                                                       & \textbf{2.68}                        & \textbf{0.886}                       & 0.951\\                         
 \hline
\end{tabular}
\vspace{-2em}
\end{table}

\noindent\paragraph{FAUST \& SHREC19} In the last task, we follow the official repository of HCLV2S~\cite{huang20} to sample 100 partial view point clouds from each shape in the FAUST and SHREC19 dataset, resulting respectively $10000$ and $4400$ test cases. It is worth noting that while shapes in FAUST share the same triangulation with the template (as used in~\cite{huang20}), SHREC19 contains shapes of much more significant variability in pose, style and vertex numbers. 

As illustrated in Tab.~\ref{table:hcv2s}, our method significantly surpasses HCLV2S, achieving over 20$\%$ and 40$\%$ relative error reduction on the FAUST and SHREC19 datasets, respectively. This further confirm the robustness and efficiency of our method when applied to partial-view datasets.

\subsubsection{Matching Alien Point Clouds}
In this section, we use \textbf{Ours-M\&P} throughout. 
We demonstrate the robustness of our model to heterogeneous partial point clouds, noisy real scans with outlier points, and capacity of matching with unseen source shape during training. 

\begin{figure*}
  \centering
  \begin{subfigure}{0.32\linewidth}
    \includegraphics[width=\linewidth]{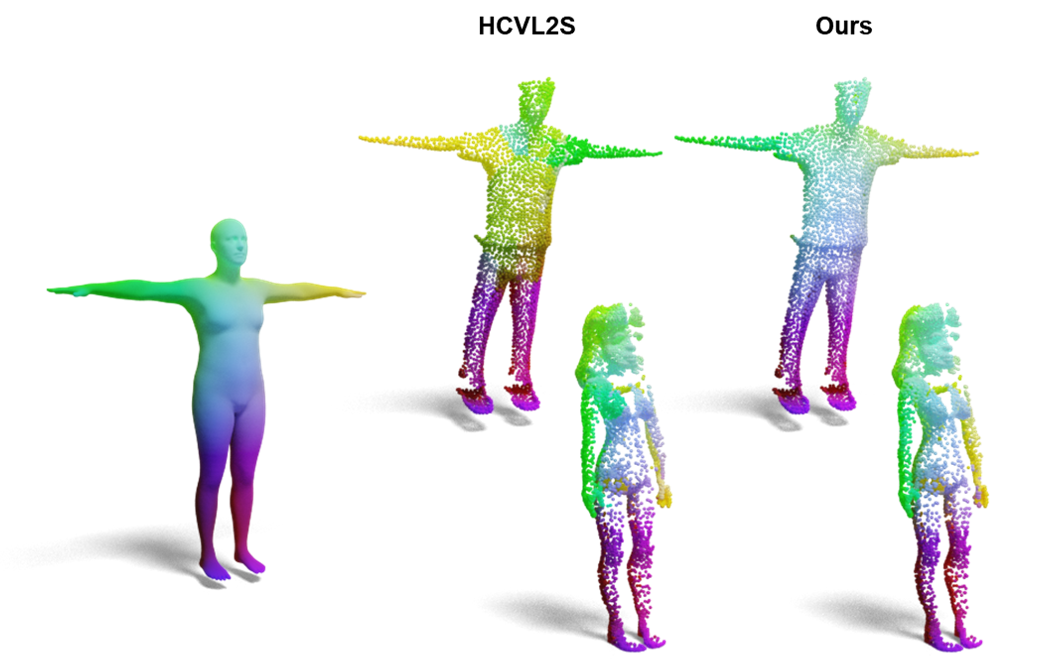}
    \caption{Heterogeneous partial point clouds.}
    \label{fig:shrec07}
  \end{subfigure}
  \hfill
  \begin{subfigure}{0.32\linewidth}
    \includegraphics[width=\linewidth]{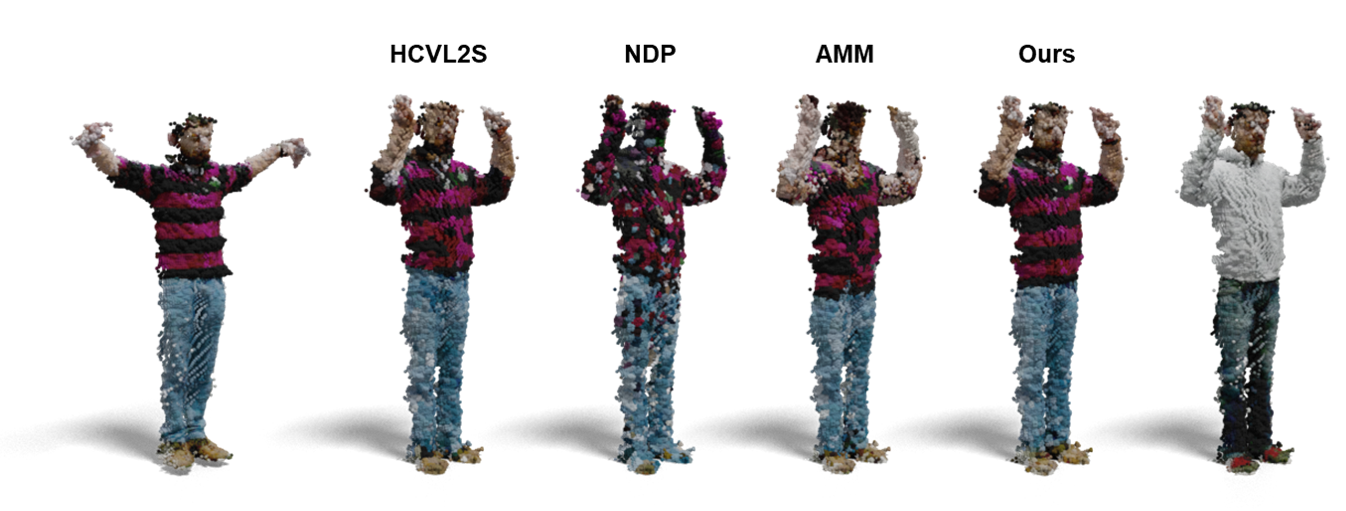}
    \caption{Noisy real scans.}\label{fig:realscan}
  \end{subfigure}
  \hfill
  \begin{subfigure}{0.32\linewidth}
    \includegraphics[width=\linewidth]{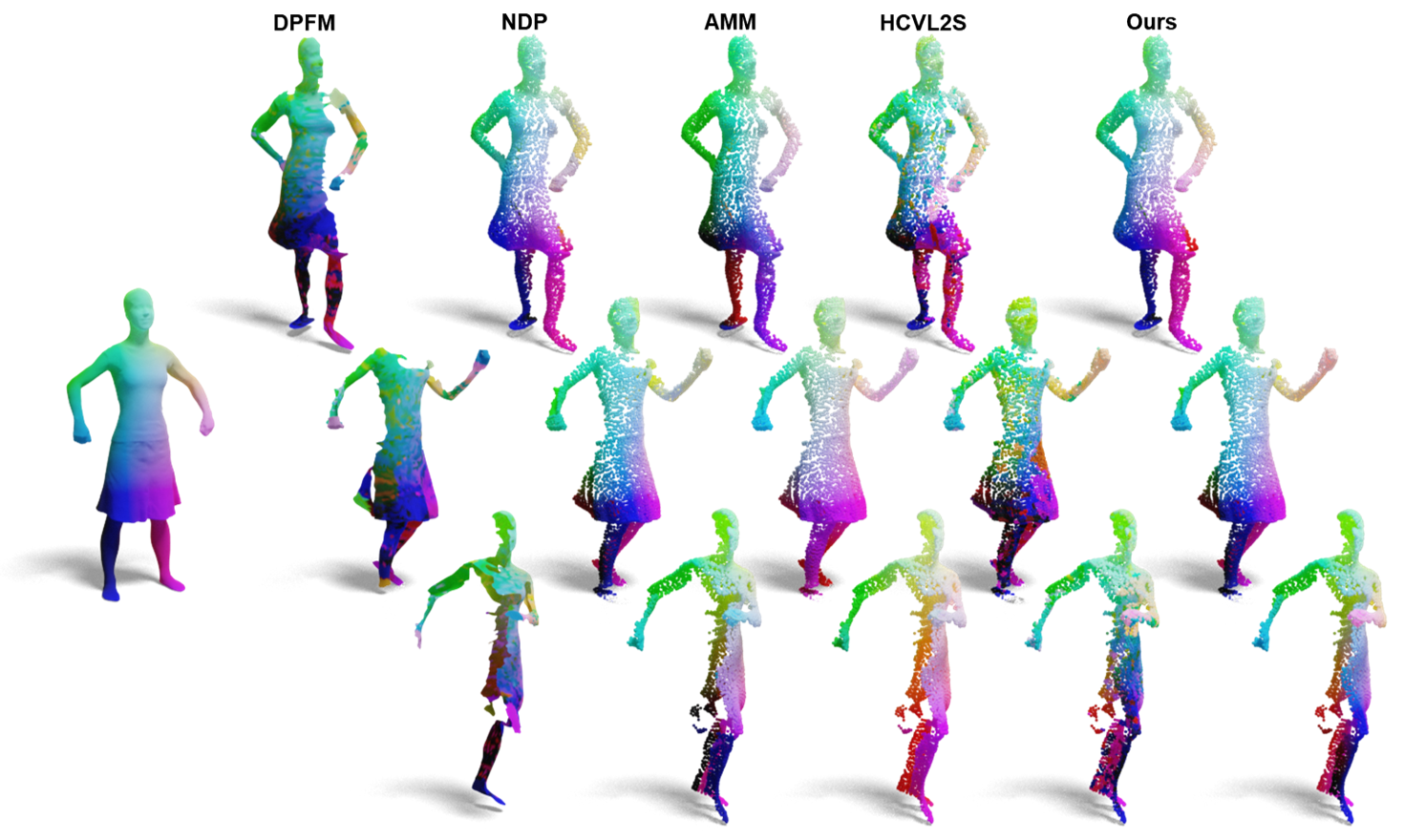}
    \caption{Different source shape selection.}
    \label{fig:mit}
  \end{subfigure}
  \caption{Qualitative comparison with different baselines.}
  \label{fig:pv}
\end{figure*}
To start with, as shown in Fig.~\ref{fig:shrec07} ,we compare with HCLV2S~\cite{huang20} on matching heterogeneous partial point clouds with the template from HCLV2S. 
It is evident from the top row that the competing method can not handle alien body type (the male shape is dressed, and differs from most of training shape from SURREAL used in HCLV2S). 

In the second set of experiment illustrated in Fig. ~\ref{fig:realscan}, apart from HCLV2S, we also compare with non-rigid shape registration methods. 
Similar to the last example, we keep using the template in Fig.~\ref{fig:shrec07} as source shape. 
Visual inspection clearly shows that our texture transfer outperforms all the baselines by a significant margin. 
Note, especially, how the facial details, strips in the T-shirt and arms are transferred by different methods. 
We remark that our texture transfer is purely geometric, without using any texture information in the input. 

In the last example, we showcase the flexibility of our pipeline in terms of source shape selection. 
In contrast, HCLV2S~\cite{huang20} is trained to match any input with a fixed template shape. 
In the case of Fig.~\ref{fig:mit}, the input point clouds are sampled from a sequence of girl in a robe dancing samba, which significantly deviates from the fixed template from SURREAL, thus leading to degraded output as shown in the right-most but second column. 
Shape registration methods~\cite{li2022non} fail due to the relative large deformation undergoing between the source and the target point clouds. 
Finally, we train DPFM~\cite{attaiki2021dpfm} on the training set of \textbf{S\&F} dataset, and test it on the largest connected component of the input mesh (which none of the rest uses), yet it still fail to deliver reasonable results. 


\subsubsection{Extended Evaluation}\label{sec:extended}

\noindent\paragraph{Partial-to-Partial Matching and Animal Shapes}
To further assess the generalization capacity of our pipeline, we evaluate on additional settings as shown in Tab.~\ref{table:data}.
First, we evaluate on partial-to-full matching on DT4D-H, where our method achieves $4.05$ \emph{vs.} $7.23$ by HCLV2S, demonstrating strong generalization to unseen humanoid shapes.
For \emph{partial-to-partial} matching, we compose two partial-to-full registrations via a shared template mesh. Our pipeline achieves $2.96$ \emph{vs.}~$8.10$ by HCLV2S, validating its effectiveness even in this more challenging setting.
Finally, we test on the TOSCA animal dataset~\cite{tosca} (trained on SMAL~\cite{zuffi20173d}), where our method reduces the error from $26.37$ (HCLV2S) to $9.78$, confirming the applicability of our framework beyond human shapes.
\begin{table}[!t]
    \caption{Mean geodesic errors ($\times$100) on extended evaluation settings: humanoid shapes (DT4D-H, partial-to-full), partial-to-partial matching (S\&F-PV), and animal shapes (TOSCA, trained on SMAL).}\label{table:data}
    \centering
    \scriptsize
    \begin{tabular}{lccc}
        \hline
        Geo.~error($\times$100) & DT4D-H & Partial2Partial & TOSCA \\ \hline
        HCLV2S~\cite{huang20}  & 7.23 & 8.10 & 26.37 \\
        \rowcolor[HTML]{E7E6E6}
        Ours  & \textbf{4.05} & \textbf{2.96} & \textbf{9.78} \\ \hline
    \end{tabular}
    \vspace{-1em}
\end{table}

\noindent\paragraph{Comparison with Pointwise Features}
To isolate the contribution of our neural feature guidance, we compare with different feature extractors on the S\&F-PV dataset in Tab.~\ref{table:feature_comp}.
We report both Stage-I output (w/o registration) and the full pipeline output (w/ registration).
Our feature extractor significantly outperforms DFR~\cite{dfr} ($4.41$ vs. $6.75$ without registration, $2.33$ vs. $3.49$ with registration), SSMSM~\cite{cao2023} ($4.41$ vs. $8.37$), and DiffFMaps~\cite{lie} ($4.41$ vs. $12.04$).
Notably, DiffFMaps~\cite{lie} \emph{degrades} with registration ($15.4$ vs. $12.04$), indicating that low-quality features can misguide the optimizer.
We additionally evaluated axiomatic pointwise descriptors including HKS~\cite{sun2009concise} ($73.73$), WKS~\cite{aubry11} ($49.28$), and SHOT~\cite{tombari2010unique} ($32.73$); these classical features lack the semantic discriminability needed for large deformations and cannot benefit from the registration pipeline.
\begin{table}[!t]
    \caption{Mean geodesic errors ($\times$100) comparing different feature extractors on S\&F-PV dataset, with and without registration refinement.}\label{table:feature_comp}
    \centering
    \scriptsize
    \begin{tabular}{lcc}
        \hline
        Geo.~error($\times$100) & w/o reg. & w/ reg. \\ \hline
        DFR~\cite{dfr}   & 6.75 & 3.49 \\
        SSMSM~\cite{cao2023} & 8.37 & 4.02 \\
        DiffFMaps~\cite{lie} & 12.04 & 15.4 \\
        \rowcolor[HTML]{E7E6E6}
        Ours  & \textbf{4.41} & \textbf{2.33} \\ \hline
    \end{tabular}
    \vspace{-1em}
\end{table}

\subsubsection{Ablation Study}\label{sec:abl}

We report ablation studies in Tab.~\ref{table:icosahedron}, in which we verify the effectiveness of core designs formulated in Sec.~\ref{sec:method}. Throughout this part, we train on \textbf{Ours-S\&F} and test on it.

As demonstrated in Tab.~\ref{table:icosahedron}, utilizing the dense full-to-full map directly as supervision, in place of the functional map, results in significant geodesic errors. Particularly, our method outperforms DFR~\cite{dfr} over 30$\%$ in both stages ($\mathbf{2.33}$ vs. $3.49$, $\mathbf{4.41}$ vs. $6.75$). We note that ``Ours feature'' and ``DFR feature'' in Tab.~\ref{table:icosahedron} correspond to the Stage-I output only (\emph{i.e.,} correspondences estimated from the learned features \emph{without} Stage-II coordinate-based refinement), thereby isolating the feature extractor quality from the registration optimizer.

To further disentangle the contributions of the feature extractor and the optimizer, we compare DFR and Partial-DFR on external benchmarks in Tab.~\ref{table:dfr}. On the PFARM dataset, DFR's feature extractor yields a geodesic error of $47.29$ (vs. ours $18.41$), and even after registration, DFR achieves only $32.87$ (vs. ours $\mathbf{7.11}$). This confirms that the performance gain stems from \emph{both} the improved Partial-DFR feature extractor \emph{and} the unidirectional CD-based registration, with the former contributing the larger share.
\begin{table}[!t]
    \caption{Mean geodesic errors ($\times$100) comparing DFR and Partial-DFR (Ours) on partial matching benchmarks, with and without registration. This isolates the contributions of the feature extractor from the registration optimizer.}\label{table:dfr}
    \centering
    \scriptsize
    \setlength{\tabcolsep}{1.2mm}
    \begin{tabular}{lcccc}
        \hline
        \multirow{2}{*}{} & \multicolumn{2}{c}{PFARM} & \multicolumn{2}{c}{SURREAL-PV} \\ \cline{2-5}
             & w/o reg. & w/ reg. & w/o reg. & w/ reg. \\ \hline
        DFR~\cite{dfr}  & 47.29 & 32.87 & 13.5 & 5.77 \\
        \rowcolor[HTML]{E7E6E6}
        Ours & \textbf{18.41} & \textbf{7.11} & \textbf{8.13} & \textbf{4.9} \\ \hline
    \end{tabular}
    \vspace{-1em}
\end{table}

\begin{table*}[]
    \caption{Statistical shape analysis on spleen and pancreas medical dataset in terms of chamfer distance. The \textbf{best} is highlighted.}\label{table:med}
    \centering
    \scriptsize
    \setlength{\tabcolsep}{0.4mm}
    \begin{tabular}{cccccccc}
    \hline
    \textbf{CD (mm)}   & PN-AE~\cite{achlioptas2017latent_pc}   & DG-AE~\cite{dgcnn}                & CPAE~\cite{cheng2021learning}   & ISR~\cite{chen2020unsupervised}                 & DPC~\cite{lang2021dpc}                  & Point2SSM~\cite{adams2023point2ssm}            & \cellcolor[HTML]{E7E6E6}Ours         \\ \hline
    \textbf{Spleen} & 43.7                 & 43.5                 & 61.3                 & 17.6                 & 10.6                 & 3.4                  & \cellcolor[HTML]{E7E6E6}\textbf{1.7} \\ 
    \textbf{Pancreas} & 22.0                 & 21.0                 & 18.8                 & 7.4                 & 6.1                & 2.7                 & \cellcolor[HTML]{E7E6E6}\textbf{1.3} \\\hline
            & \multicolumn{1}{l}{} & \multicolumn{1}{l}{} & \multicolumn{1}{l}{} & \multicolumn{1}{l}{} & \multicolumn{1}{l}{} & \multicolumn{1}{l}{} & \multicolumn{1}{l}{}                
    \end{tabular}
    \vspace{-1.5em}
\end{table*}
\begin{table}[!ht]
    \caption{Mean geodesic errors (×100) on SCAPE-FAUST dataset. The \textbf{best} is highlighted.}
    \centering
    \scriptsize
    \begin{tabular}{cccc}
        \toprule
        Method  & Geo. error(×100) \\
        \hline
        \rowcolor[HTML]{E7E6E6} 
        Ours full	&\textbf{2.33}	\\
        Ours feature	&4.41	\\

        Ours w/o $\Phi$ in training &36.55	\\

        DFR~\cite{dfr}	&3.49	\\
        DFR~\cite{dfr} feature &6.75	\\
        \bottomrule
    \end{tabular}\label{table:icosahedron}
    \vspace{-2em}
\end{table}
\subsection{Medical Application}
To further validate the effectiveness of our approach on medical application, we normalized the stomach subset from 60 patients in MedShapeNet~\cite{li2023medshapenet} (comprising 3,540 pairs), and obtained 12 partial shapes for each shape through the perspective processing of a regular dodecahedron (consistent with the method of processing partials in Tab.~\ref{table:sf}). Based on these pairs, we conducted training for both full and partial shapes using SSMSM\cite{cao2023}, Point2SSM~\cite{adams2023point2ssm}, and our method. Our method demonstrated the best performance in stomach medical dataset registration, whether full or partial cases. Besides, following Point2SSM~\cite{adams2023point2ssm}, we test our method on the spleen and pancreas subset for quantitative results. As shown in Tab.~\ref{table:med}, our method outperforms the second best by a $\textbf{50\%}$ relative error reduction. 

\section{Conclusion}
\label{sec:conc}
In this paper, we propose a novel \emph{neural feature-guided shape registration} (NFR) framework without correspondence supervision, which is also suitable for partial shape. Our framework can perform registration between shapes undergoing significant intrinsic deformations, and exhibits superior generalizability over the learning-based competitors. Apart from several designs tailored for our intuitive pipeline, we also introduce a data-driven solution to facilitate the burden of extrinsically aligning non-rigid points. To better handle the partial-2-full cases, NFR introduces a novel scheme that enhances the robustness of partial functional maps by computing spectral embeddings of partial shapes, conditioned on their complete versions. Further, we develop a point feature extractor trained on a dataset of complete meshes in a self-supervised manner. We verify our frameworks through a series of challenging tests, in which it shows superior performance but also remarkable robustness.

\noindent \textbf{Limitation and Future Work:} The primary limitation of our method lies in its optimization process, which involves iterative updates of correspondences between intermediate and target point clouds. This procedure requires several seconds to converge, indicating opportunity for improving efficiency.

\noindent \textbf{Data availability:}  We claim to release the dataset and code upon acceptance. The datasets generated and analyzed during the current study will be available in our open-source repository.

\noindent \textbf{Ethics statement:} Ethical approval was not required as the study involved neither human nor animal subjects.

\noindent \textbf{Acknowledgement:} This work was supported by the National Natural Science Foundation of China under contract No. 62171256, 62331006.



\bibliographystyle{eg-alpha-doi}
\bibliography{egbibsample}

\clearpage
\appendix
\label{sup:appendix}


\section{Proof of Proposition 1}

We restate Proposition~1 for convenience: Let $\s, \st$ be a pair of shapes each having non-repeating Laplacian eigenvalues, which are the same (\emph{i.e.,} $\Delta_\s = \Delta_\st$), and $\Pi_{\st\s}$ be an isometry between $\st$ and $\s$. Let $\Phi_\s, \Phi_\st$ be the corresponding eigenvectors, and $\st_p$ be a sub-sampled set of vertices of $\st$ with spatially truncated embedding $\Phi_{\st_p} = \Pi_{\st_p \st}\Phi_\st$. Then $C_{\s\st} = \Phi_\st^{\dagger}\Pi_{\st\s}\Phi_\s$ satisfies:
\begin{equation*}
C_{\s\st} = \arg\min_C \|\Phi_{\st_p} C - \Pi_{\st_p\s}\Phi_\s\|_{\mbox{Fro}}^2.
\end{equation*}

\noindent\textit{Proof.}
The proof proceeds in three steps.

\noindent\textbf{Step 1: Structure of $C_{\s\st}$.}
Since $\s$ and $\st$ are isometric and share the same eigenvalues $\Delta_\s = \Delta_\st$ with no repeated entries, by Lemma A.1 of~\cite{zoomout}, the functional map $C_{\s\st} = \Phi_\st^{\dagger}\Pi_{\st\s}\Phi_\s$ must be a diagonal orthogonal matrix, \emph{i.e.,} $C_{\s\st} = \text{diag}(\sigma_1, \ldots, \sigma_k)$ with $\sigma_i \in \{+1, -1\}$ for all $i$. This means that the $i$-th eigenvector of $\Phi_\s$ and that of $\Phi_\st$ differ by at most a sign flip.

\noindent\textbf{Step 2: Global optimality for full shapes.}
Since $C_{\s\st}$ is diagonal orthogonal, we have $\Pi_{\st\s}\Phi_\s = \Phi_\st C_{\s\st}$, which implies:
\begin{equation*}
    \|\Phi_\st C_{\s\st} - \Pi_{\st\s}\Phi_\s\|_{\mbox{Fro}}^2 = 0.
\end{equation*}
As the Frobenius norm is non-negative, $C_{\s\st}$ is the \textbf{global} minimizer of $\|\Phi_\st C - \Pi_{\st\s}\Phi_\s\|_{\mbox{Fro}}^2$ over all $k\times k$ matrices $C$.

\noindent\textbf{Step 3: Extension to partial shapes.}
Substituting the spatially truncated embedding $\Phi_{\st_p} = \Pi_{\st_p\st}\Phi_\st$ and the composed map $\Pi_{\st_p\s} = \Pi_{\st_p\st}\Pi_{\st\s}$, we have:
\begin{equation*}
\begin{aligned}
    \|\Phi_{\st_p} C_{\s\st} - \Pi_{\st_p\s}\Phi_\s\|_{\mbox{Fro}}^2 &= \|\Pi_{\st_p\st}\Phi_\st C_{\s\st} - \Pi_{\st_p\st}\Pi_{\st\s}\Phi_\s\|_{\mbox{Fro}}^2 \\
    &= \|\Pi_{\st_p\st}(\Phi_\st C_{\s\st} - \Pi_{\st\s}\Phi_\s)\|_{\mbox{Fro}}^2 = 0,
\end{aligned}
\end{equation*}
where the last equality follows from Step 2. Since this objective is non-negative for any $C$ and equals zero at $C = C_{\s\st}$, we conclude that $C_{\s\st}$ is the global minimizer of $\|\Phi_{\st_p} C - \Pi_{\st_p\s}\Phi_\s\|_{\mbox{Fro}}^2$, \emph{regardless} of the choice of subsampling $\st_p \subset \st$. $\square$

\noindent\textbf{Remark.}
In practice, shapes are only approximately isometric, and eigenvalues may have near-multiplicities. In such cases, $C_{\s\st}$ is approximately diagonal, and Proposition~1 holds in an approximate sense: the full-to-full functional map remains a good alignment of the truncated spectral embedding, with the approximation error controlled by the degree of non-isometry and the spectral gap.

\section{Derivation of Closed-Form Solution to Loss in Eqn.(7)}

Let $\frac{\partial}{\partial C } C_{\mbox{opt}} = 0$ then we have:

\begin{equation}
\begin{split}
\frac{\partial}{\partial C } (&\left\|\Phi_{\st_p} C-\Pi_{\st_p\s} \Phi_{\s}\right\|_F^2 +\lambda\left\|\Delta_{\st} C-C \Delta_{\s}\right\|_F^2)\\
&=2 \Phi_{\st_p}^T(\Phi_{\st_p} C-\Pi_{\st_p\s} \Phi_{\s}) + 2\lambda \Delta \cdot C = 0,
\end{split}
\end{equation}
where the operation $\cdot$ represents the element-wise multiplication, and $\Delta _{i j}=\left(\mu_j^{ \st }-\mu_i^{ \s }\right)^2$, where $\mu_l^{ \st }$ and $\mu_l^{ \s }$ respectively correspond to the eigenvalues of $\Delta_{ \st }$ and $\Delta_{ \s }$. Inspired by ~\cite{donati2020deep}, for every row $c_i$ of $\mathbf{C}$:
\begin{align}
\left(  \Phi_{\st_p}^T\Phi_{\st_p} +\lambda \operatorname{diag}\left(\left(\mu_j^1-\mu_i^2\right)^2\right)\right) c_i=\Phi_{\st_p}^T b_i,
\end{align}
where $b_i$ stands for $i^{t h}$ row of $\Pi_{\st_p\s} \Phi_{\s}$. Since solving a linear system is differentiable in Pytorch, this allows us to estimate the
functional map during training.

\section{Technical Details}\label{sup:B}
\subsection{Orientation Regressor}\label{sup:or}

Orientation plays an important role in point cloud processing. Classic point-based feature extractors~\cite{qi2017pointnet, 2017Qipointnet2, dgcnn} are not rotation-invariant/equivariant by design and, therefore are limited. A common practice is to leverage data augmentation~\cite{qi2017pointnet}. However, such can only robustify the model under small perturbations, while falling short of handling arbitrary rotations in $SO(3)$. On the other hand, recent advances in designing rotation-invariant/equivariant feature extractor~\cite{poulenard2019SPHNet, deng2021vn} seem to shed light on alleviating alignment requirements in non-rigid point matching. However, we empirically observe that such a design generally degrades the expressive power of the feature extractor, leading to inferior performance of the trained DFM.

This issue also attracts attention from the community of shape matching. In~\cite{deng2023se}, the rotation prediction, instead of directly estimating the transformation, is formulated as a classification problem. However, this method can only address the rotation around a fixed axis and exhibits relatively low accuracy. Other works~\cite{yu2022riga, yu2023rotation} transform the input point cloud into handcrafted features. However, it should be noted that this approach often increases the computational complexity due to the additional feature extraction and processing steps involved.

In this paper, we propose a data-driven solution to this problem, accompanied by a registration module that is robust with the estimated alignment. We adopt the regressor model proposed in~\cite{chen2022projective}. We use the synthetic SURREAL shapes~\cite{varol17_surreal}, which are implicitly aligned by the corresponding generative codes to train our orientation regressor. We follow ~\cite{chen2022projective} and use PointNet++~\cite{2017Qipointnet2} as the feature extractor.
As we assume to be given full shape, either mesh or point cloud, for each input cloud, we translate it so its mass lies at the origin point. Then we estimate the orientation deviation, $R$, of it from the orientation implicitly defined by the SURREAL dataset. Finally, we apply the inverse rotation $R^T$ on the input to obtain the aligned shape. We emphasize again that all the shapes, either train or test, go through this alignment in our pipeline.

\subsection{Addition of DFR Training}
 \noindent\textbf{Training DFM on Meshes: } Our training scheme in principle follows that of~\cite{dual}, which delivers both accurate map estimation and excellent generalization performance.

 1) We take a pair of shapes $\src_1, \src_2$, and compute the leading $k$ eigenfunctions of the Laplace-Beltrami operator on each shape. The eigenfunctions are stored as matrices $\Phi_i \in \mathbb{R}^{n_i \times k}, i = 1, 2$.

 2) We compute $G_1 = \gm(\src_1), G_2 = \gm(\src_2)$, projected into the spaces spanned by $\Phi_1, \Phi_2$ respectively, and leverage the differential FMReg module proposed in~\cite{donati20} to estimate the functional maps $C_{12}, C_{21}$ between the two shapes, then by regularizing the structure of $C_{12}, C_{21}$, say:
\begin{equation}\label{sup:bo}
    \begin{aligned}
    E_{\mbox{bij}}(\gm) &= \| C_{12} C_{21} - I\|_F^2, \\
    E_{\mbox{ortho}}(\gm) &= \| C_{12}C_{12}^T - I\| + \| C_{21}C_{21}^T - I\|.
    \end{aligned}
\end{equation}

 3) To enhance $\gm$, we estimate the correspondences between $\src_1, \src_2$ via the proximity among the rows of $G_1, G_2$ respectively. Namely, we compute
 \begin{equation}\label{sup:pi}
\Pi_{12}(i, j) = \frac{\exp(-{\alpha\delta_{ij}})}{\sum_{j'} \exp(-\alpha\delta_{ij'})}, \forall i\in [|\src_1|], j\in [|\src_2|],
\end{equation}
where $\delta_{ij} = \|G_1(i, :) - G_2(j, :)\|$ and $\alpha$ is the temperature parameter, which is increased during training~\cite{dual}. Then $\Pi_{12}$ is the soft map between the two shapes. Ideally, it should be consistent with the functional maps estimated in Eqn.(\ref{sup:bo}), giving rise to the following loss:
\begin{equation}\label{sup:cc}
 E_{\mbox{align}}(\gm) = \|C_{12} - \Phi_2^{\dagger} \Pi_{21} \Phi_1\| + \|C_{21} -  \Phi_1^{\dagger} \Pi_{12} \Phi_2\|,
\end{equation}
where $\dagger$ denotes to the pseudo-inverse. To sum up, $\gm^* = \arg\min E_{\mbox{DFM}}(\gm)$, which is defined as:
\begin{equation}\label{sup:dfm1}
    \begin{aligned}
	&E_{\mbox{DFM}}(\gm) = \lambda_{\mbox{bij}}E_{\mbox{\mbox{bij}}}(\gm) + \lambda_{\mbox{orth}}E_{\mbox{orth}}(\gm)
    \\& + \lambda_{\mbox{align}}E_{\mbox{align}}(\gm).
    \end{aligned}
\end{equation}

\noindent\textbf{Training DFM on Point Clouds: }Now we are ready for training $\fm$ with a modified DGCNN~\cite{nie} as backbone. We let $F_1 = \textbf{F}(\src_1), F_2 = \textbf{F}(\src_2)$. On top of $E_{\mbox{DFM}}$ in Eqn.~(\ref{sup:dfm1}), we further enforce the extracted feature from $\fm$ to be aligned with that from $\textbf{G}^*$ via PointInfoNCE loss~\cite{xie2020pointcontrast}:
\begin{equation}\label{sup:nce}\
    \begin{aligned}
	&E_{\mbox{NCE}}(\fm, \gm^*) = -\sum_{i=1}\limits^{n_1} \log\frac{\exp(\langle F_1(i, :),G^*_1(i, :)\rangle/\gamma)}{\sum_{j=1}\limits^{n_1} \exp(\langle F_1(i, :), G^*_1(j, :)\rangle/\gamma)}\\&-
 \sum_{i=1}\limits^{n_2} \log\frac{\exp(\langle F_2(i, :),G^*_2(i, :)\rangle/\gamma)}{\sum_{j=1}\limits^{n_2} \exp(\langle F_2(i, :), G^*_2(j, :)\rangle/\gamma)},
    \end{aligned}
\end{equation}
where $\langle.\rangle$ refers to the dot products, $\gamma$ is the temperature parameter, and $n_i$ is the number of points of $\src_i$. So the final training loss is defined as:
\begin{equation}\label{sup:dfm}
	E(\fm) = E_{\mbox{DFM}}(\fm) + \lambda_{\mbox{NCE}}E_{\mbox{NCE}}(\fm, \gm^*),
\end{equation}
which is optimized on all pairs of shapes within the training set.

\subsubsection{Shape Registration}\label{sup:reg}
Now we are ready to formulate our registration module. We first describe the deformation graph construction in our pipeline.
Then we formulate the involved energy functions and finally describe the iterative optimization algorithm. During  registration, we denote the deformed source model by $\mathcal{S}^{k} = \{ \mathcal{V}^{k}, \mathcal{E} \}$,$ \mathcal{V}^{k}=\left\{ v_i^k \mid i=1, \ldots, N\right\}$, where $k$ indicates the iteration index and $v_i^k$ is the position of the $i-$th vertex at iteration $k$. The target
point cloud is denoted by $\mathcal{T} =\left\{ u_j \mid j=1, \ldots, M\right\}$.

\noindent\textbf{Deformation Graph: } Following\cite{guo2021human}, we reduce the vertex number on $\src$ to $H = [N/2]$  with QSlim algorithm~\cite{garland1997surface}. Then an embedded deformation graph $\mathcal{DG}$ is parameterized with axis angles $\Theta\in \mathbb{R}^{H \times 3}$ and translations $\Delta\in \mathbb{R}^{H \times 3}$: $\textbf{X} = \{\Theta , \Delta  \}$. The vertex displacements are then computed from the corresponding deformation nodes. For a given $\textbf{X}^{k}$, we can compute displaced vertices via:
 \begin{equation}\label{sup:dg}
     \mathcal{V}^{k+1} = \mathcal{DG}(\textbf{X}^{k},\mathcal{V}^{k}).
 \end{equation}

\noindent\textbf{Dynamic correspondence update: } Thanks to our point-based feature extractor $\fm$, we can freely update correspondences between deforming source mesh and target point clouds. In practice, for the sake of efficiency, we update every 100 iterations in optimization (see Alg. 1 in the main text).

\noindent\textbf{Correspondence Filtering via Bijectivity: } One key step in shape registration is to update correspondences between the deforming source and the target. It is then critical to reject erroneous corresponding points to prevent error accumulation over iterations. In particular, we propose a filter based on the well-known bijectivity prior. Given $\src^k$ and $\tar$, we first compute maps represented as permutation matrices $\Pi_{\src\tar}, \Pi_{\tar\src}$, either based on the learned feature or the coordinates. For $v_i^k\in \src^k$, we compute the geodesic on $\src^k$ between $v_i^k$ and its image under permutation $\Pi_{\src\tar}\Pi_{\tar\src}$, then reject all $v_i^k$ if the distance is above some threshold. For the sake of efficiency, we pre-compute the geodesic matrix of $\src$ and approximate that of $\src^k$ with it. Finally, we denote the filtered set of correspondences by $\mathcal{C}^k = \{(v_{i_1}^k, u_{j_1}), (v_{i_2}^k, u_{j_2})\cdots\}$.

In the following, we introduce the energy terms regarding our registration pipeline.

\noindent\textbf{Correspondence Term} measures the distance of filtered correspondences between $\src^k$ and $\tar$, given as:
\begin{equation}
      E_{\mbox{corr}} = \frac{1}{|\mathcal{C}^k|} \sum_{(v_i^k,u_j) \in \mathcal{C}^k }\left\| v^{k}_{i} - u_{j}\right\|_2^2.
\end{equation}

\noindent\textbf{Chamfer Distance Term} has been widely used~\cite{li2022non, li2022lepard} to measure the extrinsic distance between $\src^k$ and $\tar$:

\begin{equation}
    \begin{aligned}
    &E_{\mbox{cd}}=\frac{1}{N} \sum_{ i \in [N] } \min _{ j \in [M] } \left\| v^{k}_{i} - u_{j}\right\|_2^2
    \\&+\frac{1}{M} \sum_{ j \in [M]} \min_{ i \in [N] } \left\| v^{k}_{i} - u_{j}     \right\|_2^2
    \end{aligned}
\end{equation}

\noindent\textbf{As-rigid-as-possible Term} reflects the deviation of estimated local surface deformations from rigid transformations. We follow~\cite{guo2021human, levi2014smooth} and define it as:

\begin{equation}
    E_{\mbox{arap}} =\sum_{h \in [H]} \sum_{l \in \psi(h)}(\left\|d_{h, l}( \textbf{X})\right\|_2^2  + \alpha \left\|(R\left( \Theta_h\right)-R\left( \Theta_l \right)\right\|_2^2 )
\end{equation}

 Here, $g \in R^{H\times 3}$ are the original positions of the nodes in the deformation graph $\mathcal{DG}$, and $\psi(h)$ is the 1-ring neighborhood of the $h-$th deformation node. $R(\cdot)$ is Rodrigues' rotation formula that computes a rotation matrix from an axis-angle representation and $\alpha$ is the weight of the smooth rotation regularization term.

\section{Robustness}
We evaluate the robustness of models trained on $S\&F$ dataset with respect to noise and rotation perturbation and report in Table~\ref{table:robust} (partial-2-full), while also evaluate on full-2-full registration in Table~\ref{table:robust_full}.

More specifically, we perturb the point clouds by: (1) Adding per-point Gaussian noise with i.i.d $\mathcal{N}(0, 0.02)$ along the normal direction on each point; (2) Randomly rotating around the gravity axis by $\pm 15$ degree.
We perform $3$ rounds of test, and report both mean error and the standard deviation in parentheses.

Note that DPFM is a mesh-based method, which enjoys rotation invariance.
Apart from that, our pipeline delivers the most robust performance among the baselines in this test.


\begin{table}[]
\caption{Mean geodesic errors (×100) on SCAPE-FAUST dataset under different perturbations. Noisy PC means the input point clouds are perturbed by Gaussian noise.  Rotated PC means the input point clouds are randomly rotated within ±15 degrees along x, y, z axises respectively. The standard deviation value is shown in parentheses.}
\centering
\scriptsize
\begin{tabular}{cccc}
\hline
\rowcolor[HTML]{FFFFFF} 
\multicolumn{1}{l}{\cellcolor[HTML]{FFFFFF}} & Unperturbed PC           & Noisy PC     & Rotated PC     \\ \hline
DPFM ~\cite{attaiki2021dpfm} unsup & 13.15 & 21.1(4.72) & - \\
NIE~\cite{nie} & 11.17 & 12.8(0.52) & 17.4(2.18) \\
Diffmaps~\cite{lie} & 12.04 & 15.3(2.88) & 22.8(4.07) \\
\rowcolor[HTML]{E7E6E6} 
ours-S$\&$F 	&\textbf{2.33}	&\textbf{2.86(0.17)}	&\textbf{4.49(0.87)}  \\ \hline
\end{tabular}\label{table:robust}
\end{table}
\begin{table}[]
\caption{Mean geodesic errors (×100) on under different perturbations. Noisy PC means the input point clouds are perturbed by Gaussian noise. Rotated PC means the input point clouds are randomly rotated within $\pm30$ degrees along $x, y, z-$axises respectively. The standard deviation value is shown in parentheses.}\label{table:robust_full}
\centering
\scriptsize
\begin{tabular}{cccc}
\hline
\rowcolor[HTML]{FFFFFF} 
\multicolumn{1}{l}{\cellcolor[HTML]{FFFFFF}} & Unperturbed PC           & Noisy PC     & Rotated PC     \\ \hline
\rowcolor[HTML]{FFFFFF} 
DiffFMaps~\cite{lie}                                    & 12.0         & 14.9(2.57)         & 26.5(3.35)         \\
\rowcolor[HTML]{FFFFFF} 
NIE~\cite{nie}                                          & 11.0         & 11.5(0.32)         & 19.9(1.29)         \\
\rowcolor[HTML]{FFFFFF} 
SSMSM~\cite{cao2023}                                        & 4.1          & 5.4(0.11)          & 9.2(1.01)          \\
\rowcolor[HTML]{E7E6E6} 
Ours                                         & \textbf{2.6} & \textbf{2.9(0.06)} & \textbf{3.6(0.96)} \\ \hline
\end{tabular}
\end{table}
\begin{table}[]
    \begin{center}
    \caption{Mean geodesic errors (×100) or Euclidean distance mean errors(cm) on SURREAL-PV/MPI-FAUST/SHREC19 dataset w/o register. Note: * means the evaluation via mean geodesic errors.}
    \centering
    \scriptsize
    \setlength{\tabcolsep}{1mm}{
    \begin{tabular}{cccc}
        \toprule
        Method  & SURREAL-PV(*) & MPI-FAUST & SHREC19\\
        \midrule
        ours-S$\&$F	&11.49&4.49&4.82	\\
        ours-M$\&$P	&8.13&2.52&4.27	\\
        \bottomrule
    \end{tabular}}\label{table:ssmsm}
    \end{center}
\end{table}

\section{Run-Time Analysis}
\begin{figure}[!htbp]
    \centering
    \includegraphics[width=0.5\textwidth]{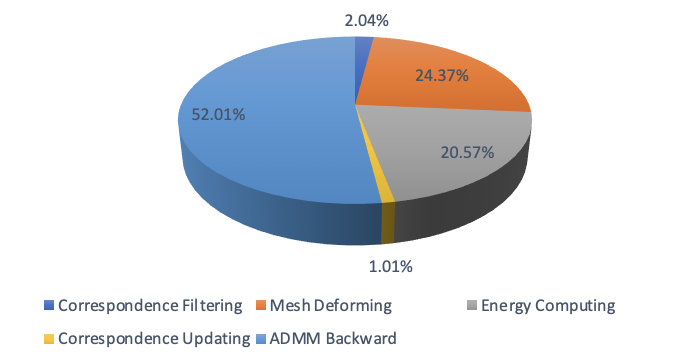}
    \caption{Run-time decomposition of our registration module.}\label{fig:time}
\end{figure}

In the last part, we provide a decomposed running time analysis on our pipeline.
In particular, we report the average running time in the test set of $\textbf{S\&F}$ dataset in Table~\ref{table:time}, as well as a time composition visualization in Figure~\ref{fig:time}. Though the reported run-time is not quite satisfying, we emphasize that to evaluate maps across $n$ point clouds, we only need to register them to the same template shape ($O(n)$), which is different from the pair-wise ($O(n^2)$) mapping and post-processing approaches~\cite{cao2022}.

\begin{table}[]
\caption{Average time and iteration cost for each shape.}\label{table:time}
\centering
\scriptsize
\setlength{\tabcolsep}{1.8mm}{
\begin{tabular}{lccc}
\hline
\qquad  & Total & Phase-I & Phase-II \\
\midrule
Time Cost(s) & 17.34 & 13.38 & 3.96 \\
Iterations & 2267 & 1757 & 510 \\
\hline
\end{tabular}}
\end{table}

\section{Scalability}\label{sup:sca}
\begin{figure}[!t]
  \begin{center}
    \includegraphics[width=0.4\textwidth]{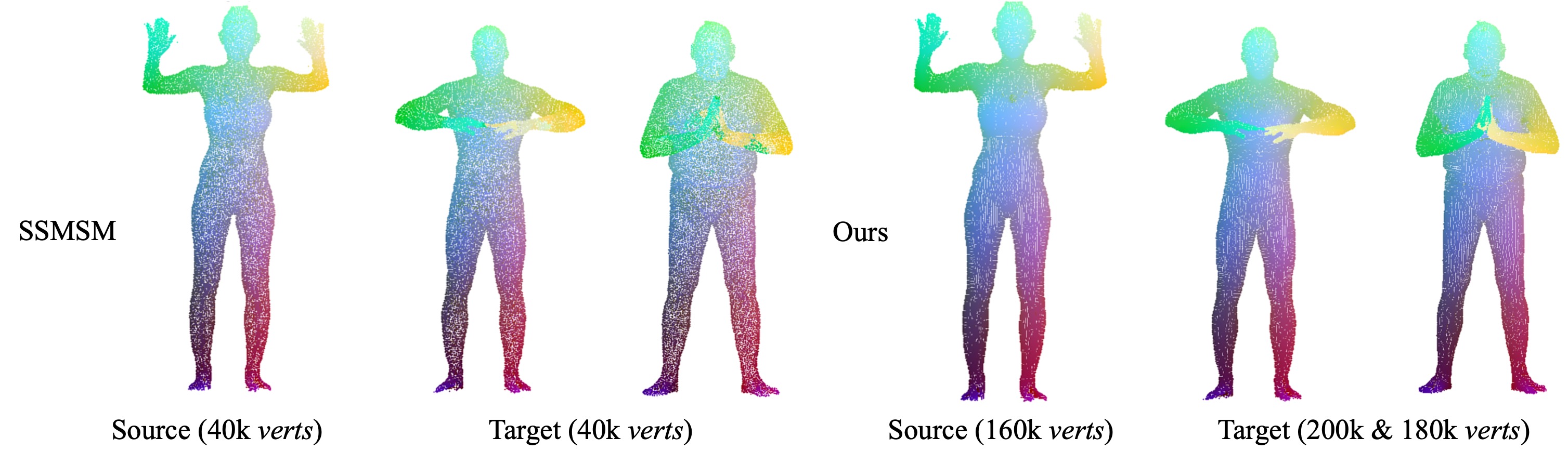}
    \end{center}
\caption{Qualitative results on matching MPI-FAUST raw scan data. Our method can match the point clouds at the original resolution, while SSMSM~\cite{cao2023} can not.  }
\label{fig:faust_mpi}
\end{figure}

Our approach is scalable w.r.t input size. Thanks to the fact that we non-rigidly align shapes in $\mathbb{R}^3$, we can in theory freely down- and up-sample both the template mesh and the target point clouds. Note this is non-trivial for methods based on mesh or graph Laplacian, as deducing dense maps with landmark correspondences over graph structures is a difficult task on its own. In Fig.~\ref{fig:faust_mpi}, we show the matching results on the real scans from the FAUST challenge, each of which consists of around 160,000 vertices. In contrast, \cite{cao2023} can handle at most 50,000 vertices without running out of 32G memory on a V100 GPU. We visualize its matching results on the subsampled (to 40,000 vertices) point clouds for comparison.

\end{document}